\documentclass[10pt,twocolumn,letterpaper]{article}

\usepackage{cvpr}      

\usepackage{algorithm}
\usepackage{algorithmic}
\usepackage{multirow}  

\usepackage{graphicx}
\usepackage{amsmath}
\usepackage{amssymb}
\usepackage{booktabs}
\usepackage[table]{xcolor} 

\usepackage{booktabs}
\usepackage{pifont}
\usepackage{xcolor}

\usepackage{pifont}
\usepackage{makecell}

\definecolor{cvprblue}{rgb}{0.21,0.49,0.74}
\usepackage[pagebackref,breaklinks,colorlinks,allcolors=cvprblue]{hyperref}

\def\paperID{*****} 
\def\confName{CVPR}
\def\confYear{2026}

\newcommand{\algoname}{CrimEdit}

\newcommand{\XY}[2]{#1_{#2}}
\newcommand{\condremoval}{\XY{c}{r}}
\newcommand{\condImage}{\XY{x}{I}}
\newcommand{\condinsertion}{\XY{c}{i}}
\newcommand{\generator}[1]{\epsilon_{\theta}#1}
\newcommand{\generatorRemoval}[1]{\epsilon_{r}}
\newcommand{\generatorInsertion}[1]{\epsilon_{i}}
\newcommand{\guidanceScale}{w}

\newcommand{\data}{\mathcal{D}}
\newcommand{\image}{I}
\newcommand{\imageObject}{\image_{o}}
\newcommand{\imageObjectAugmented}{C({\image}_{o})}

\newcommand{\imageBackground}{\image_{b}}
\newcommand{\mask}{m}
\newcommand{\pixelRegionRemoval}{\mathcal{P}_r}
\newcommand{\pixelRegionInsertion}{\mathcal{P}_i}
\definecolor{thickgreen}{RGB}{0, 150, 0}  
\definecolor{thickred}{RGB}{150, 0, 0}  

\title{\includegraphics[height=1em]{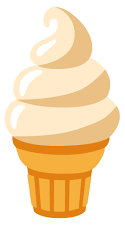} CrimEdit: Controllable Editing for Counterfactual \\ Object Removal, Insertion, and Movement}

\author{
    Boseong Jeon\textsuperscript{*},
    Junghyuk Lee\textsuperscript{*},
    Jimin Park,
    Kwanyoung Kim, \\
    Jingi Jung,
    Sangwon Lee,
    Hyunbo Shim \\
    Samsung Research \\
{\tt\small \{bf.jeon, j114.lee, jm2010.park, k\_0.kim, jingi.jung, sw96.lee, hyunbo9.shim\}@samsung.com}
}

\begin{document}
\maketitle

\renewcommand{\thefootnote}{\fnsymbol{footnote}}
\footnotetext[1]{Equal contribution}
\begin{figure*}[t!]
\centering
\includegraphics[width=0.82\linewidth]{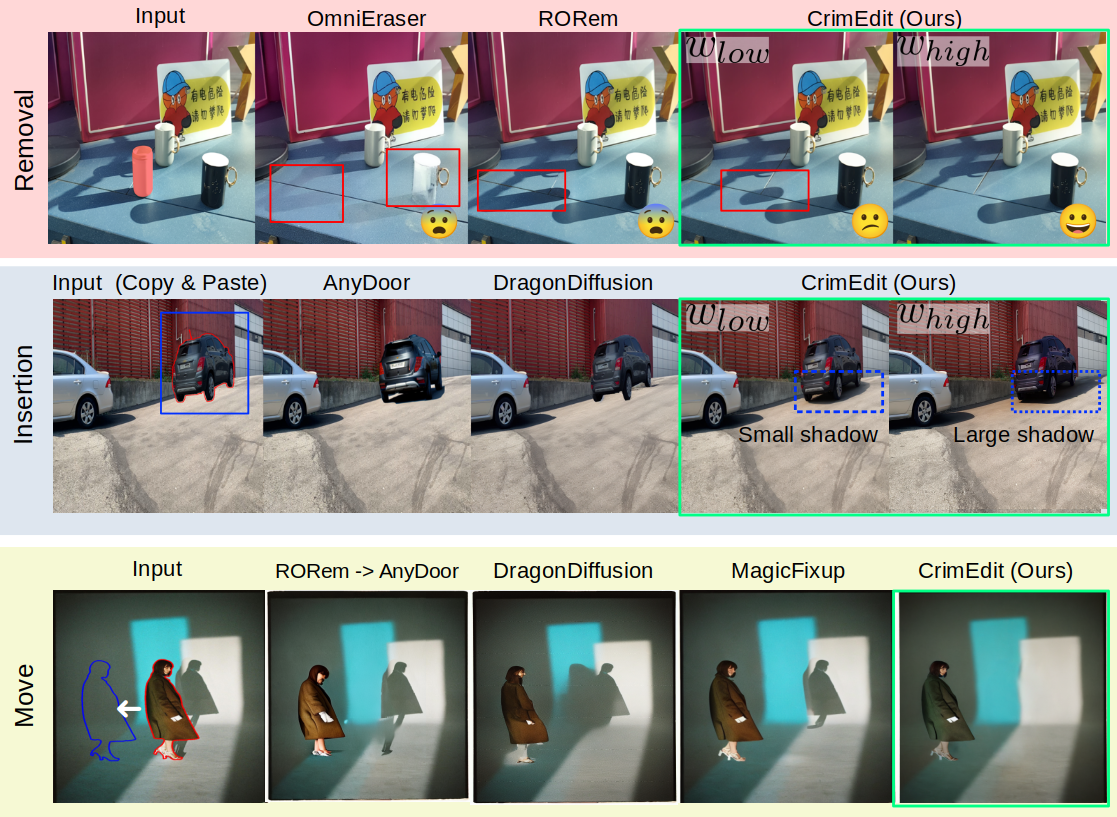}
\caption{
\algoname{} serves as a unified model to perform three tasks: object removal, insertion, and movement. 
By leveraging guidance between the task embeddings of removal and insertion, \algoname{} enhances removal quality and enables controllable effect generation for the insertion task. $w_{low}$ and $w_{high}$ denote the low and high guidance scales, respectively.
For object movement, spatial guidance allows \algoname{} to perform the task in a single denoising step—without requiring separate removal and insertion phases or additional training on a movement dataset.
}
\label{fig-thumbnail}
   
\end{figure*}

\begin{abstract}
Recent works on object removal and insertion have enhanced their performance by handling object effects such as shadows and reflections, using diffusion models trained on counterfactual datasets.
However, the performance impact of applying classifier-free guidance to handle object effects across removal and insertion tasks within a unified model remains largely unexplored.
To address this gap and improve efficiency in composite editing, we propose \textit{\algoname{}}, which jointly trains the task embeddings for removal and insertion within a single model and leverages them in a classifier-free guidance scheme—enhancing the removal of both objects and their effects, and enabling controllable synthesis of object effects during insertion.
\algoname{} also extends these two task prompts to be applied to spatially distinct regions, enabling object movement (repositioning) within a single denoising step.
By employing both guidance techniques, extensive experiments show that \algoname{} achieves superior object removal, controllable effect insertion, and efficient object movement—without requiring additional training or separate removal and insertion stages.

\end{abstract}

\section{Introduction}
Diffusion probabilistic models have recently emerged as a powerful class of generative models, achieving remarkable performance in high-quality image synthesis~\cite{sd, podell2023sdxl, dit, li2024playgroundv25insightsenhancing, cg}.
Their scalability and flexible conditioning have led to widespread adoption in computer vision and graphics.
Image editing has emerged as a particularly compelling application, offering the ability to make precise, realistic, and controllable modifications to a visual content, such as inpainting \cite{lugmayr2022repaintinpaintingusingdenoising, blendeddiffusion, xie2022smartbrushtextshapeguided, ju2024brushnet, yang2023magicremovertuningfreetextguidedimage}, object composition \cite{objectstitch, paintbyexample}, relighting \cite{jin2024neuralgafferrelightingobject, iclight}, and object repositioning \cite{seele, dragondiffusion}.

Object insertion, removal, and repositioning are fundamental image editing tasks that require not only altering image content in the target region, but also preserving consistency with the surrounding scene.
High-quality editing in these tasks depends on three key factors: 1) addressing visual effects such as shadows, lighting, and reflections introduced by objects, 2) providing adjustable editing guided by user preferences, and 3) versatility of a model for efficient deployment to users without the need to maintain multiple task-specific models.

To enable more realistic image editing, it is crucial to account for object-induced visual effects that contribute to the overall coherence of a scene. 
To this end, recent works~\cite{objectdrop, omnipaint, li2025roremtrainingrobustobject, wei2025omnieraserremoveobjectseffects} have utilized counterfactual datasets~\cite{sagong2022rord}, which contain paired examples of original and edited images with consistent scene layout and lighting. 
These datasets provide explicit supervision for learning how object presence or absence affects the surrounding visual context.

Another key aspect of image editing is controllability—specifically, the ability to adjust the intensity and characteristics of visual effects through a simple tuning process.
For example, it is desirable to let users control the size of the shadow cast when inserting an object under sunlight.
To address controllability in diffusion-based models, PowerPaint~\cite{powerpaint} introduced a prompt-learning framework and applied classifier-free guidance (CFG) between prompts to enable more directed generation.
While effective in guiding content within specified regions, this approach is limited to in-mask editing and does not account for scene-level object effects such as shadows.
The controllability and the role of task-specific guidance remain underexplored in conditional models trained with supervision on counterfactual datasets.

In practical applications, such as modern smartphone photo editors with a limited memory, users expect a seamless editing experience that supports multiple operations, such as object removal and insertion, within a single interface~\cite{samsung2024generative, apple2024iphoneguide}.
While several prior studies~\cite{objectdrop, omnipaint} have utilized counterfactual datasets to tackle both tasks, they typically rely on separate models for each.
ObjectMover~\cite{objectmover} takes a step toward unification by training a single model for both object removal and insertion, while accounting for object effects using video data.
However, ObjectMover relies on additional video footage for training, which increases data collection overhead and limits scalability. 
It also does not fully harness the potential of multi-task conditional models, such as through CFG in the conditional embeddings for fine-grained control.


To tackle the key challenges outlined above, we propose \textbf{\algoname{}}, a unified framework for \underline{C}ontrollable editing of counterfactual object \underline{R}emoval, \underline{I}nsertion, and \underline{M}ovement.
Leveraging only open-sourced counterfactual image dataset~\cite{sagong2022rord}, we first train a single diffusion model conditioned on two distinct task embeddings to perform realistic object removal and insertion within a unified architecture.
We introduce a novel guidance mechanism, \textit{Cross-Task Guidance (CTG)}, which applies classifier-free guidance between the two task embeddings. By modulating the CTG scale, our model achieves improved object removal—including associated effects such as shadows and reflections—and enables controllable synthesis of object effects during insertion.
To fully exploit task conditions trained within a single model, \algoname{} further proposes \textit{Spatial CTG (S-CTG)}, which applies CTG into spatially distinct regions. This enables simultaneous object removal and insertion in a single denoising step. Without additional training for object movement, S-CTG effectively supports object repositioning while preserving realism in object-related effects.
In summary, our main contributions are as follows:
\begin{itemize}
    \item Leveraging a network trained on both complementary tasks of object removal and insertion, we propose a novel guidance technique, \textbf{CTG}, which controls the magnitude of one task by using the conditional embeddings of the other as negative guidance.
    \item By introducing \textbf{Spatial-CTG} (S-CTG), which applies guidance to spatially distinct regions, we demonstrate that the model can perform object movement—or any task requiring simultaneous removal and insertion—in a single inference step, without the need for a dedicated training dataset.
    \item Through extensive experiments, we demonstrate that \algoname{} achieves state-of-the-art performance on object removal, insertion, and movement tasks, outperforming baselines specifically designed for each individual task. 
\end{itemize}

\section{Related Works}
This section reviews related works on object removal, insertion, and movement. A comparative summary is provided in Table~\ref{tab:method_comparison}.
\subsubsection{Object Removal}
Recent diffusion-based image inpainting models~\cite{ju2024brushnet, xie2022smartbrushtextshapeguided, blendeddiffusion, lugmayr2022repaintinpaintingusingdenoising} generate visually plausible content in missing or masked regions.
As the naive inpainting models often produce unwanted random objects inside masks,
some works have proposed removal-oriented approaches using either conditional embeddings \cite{powerpaint, ekin2024clipawayharmonizingfocusedembeddings} or inversion-based attention reweighting \cite{sun2025attentiveeraserunleashingdiffusion, yang2023magicremovertuningfreetextguidedimage}.

To enable more realistic object removal, approaches such as ObjectDrop~\cite{objectdrop}, RORem~\cite{li2025roremtrainingrobustobject}, and OmniEraser~\cite{wei2025omnieraserremoveobjectseffects} utilize counterfactual datasets~\cite{sagong2022rord}, which contain image pairs with and without the target object.
These approaches can remove not only the object itself but also its associated visual effects within the target mask region.
However, guidance techniques specifically designed for counterfactual models remain underexplored. 
Existing methods often rely on manually crafted prompts such as 'empty' or 'background' to suppress object regeneration, but such approach tends to be ineffective in general scenarios.

\subsubsection{Object Insertion}
Object insertion is to place an object from a reference image into a target image, achieving color harmonization and realistic visual effects while preserving the object's identity. 
Prior works \cite{paintbyexample, anydoor, freecompose, objectstitch} extract visual features from the reference image as conditioning inputs to the diffusion model. While these methods enable object synthesis with deformations, they often struggle to preserve fine-grained identity details and frequently produce inaccurate visual effects.
To improve the realism of synthesized effects, recent approaches \cite{objectdrop, omnipaint} leveraged counterfactual datasets and bootstrapped models originally trained for object removal.
However, these methods typically train separate models for removal and insertion, limiting their efficiency for multi-purpose editing. 
Although OmniPaint \cite{omnipaint} attempts to regulate effect synthesis via loss weighting during training, such control is not adjustable at inference time. 

\subsubsection{Object Movement}
We consider the task of object movement (repositioning), where an object and its associated effects are removed from their original location and synthesized at a new location (see the last row of Figure~\ref{fig-thumbnail}).
Several previous methods \cite{wang2024metashadowobjectcenteredshadowdetection, seele} adopt a multi-stage pipeline consisting of object removal followed by harmonized insertion.  
This approach incurs the cumulative computational cost of both stages, as high visual quality must be ensured independently in each phase. 
Drag-style editing techniques \cite{dragondiffusion, diffeditor} enable object movement in a single stage, but they require intensive computation and often suffer from background artifacts such as pixel dragging.

In contrast, FunEditor~\cite{funeditor} adopts cross-attention (CA) masking to perform the task more efficiently, applying a removal prompt to the original region and insertion prompts to the destination. This enables object movement within a single denoising step. However, it does not address object-associated visual effects. Moreover, the CA strategy is effectively equivalent to using conditional diffusion outputs without any guidance techniques such as CFG.
Other works \cite{alzayer2024magicfixup, objectmover} leverage video footage where the target object is moved across frames to naturally capture repositioning. However, collecting such datasets is expensive and introduces challenges such as unintended background changes, shifts in camera viewpoint or lighting, and deformation particularly for non-rigid targets like humans \cite{objectmover}.

\newcommand{\cmark}{{\ding{51}}} 
\newcommand{\xmark}{{\ding{55}}}           

\begin{table}[t]
\centering
\renewcommand{\arraystretch}{1.1}
\begin{tabular}{lcccccc}
\toprule
\textbf{} & \textbf{R} & \textbf{I} & \textbf{M} & \textbf{Eff.} & \makecell{\textbf{Img.} \\ \textbf{Data}} & \makecell{\textbf{Task} \\ \textbf{Guidance}} \\
\midrule
PowerPaint          & \cmark & \xmark & \xmark & \xmark & \cmark & \cmark \\
FunEditor           & \cmark & \cmark & \cmark & \xmark & \cmark & \xmark \\
\makecell[l]{RORem \& \\OmniEraser} & \cmark & \xmark & \xmark & \cmark & \cmark & \xmark \\
ObjectDrop          & \cmark & \cmark & \xmark & \cmark & \cmark & \xmark \\
OmniPaint           & \cmark & \cmark & \xmark & \cmark & \cmark & \xmark \\
ObjectMover         & \cmark & \cmark & \cmark & \cmark & \xmark & \xmark \\
\rowcolor{green!10} {\algoname{}} & \cmark & \cmark & \cmark & \cmark & \cmark & \cmark \\
\bottomrule
\end{tabular}
\caption{Comparison of capabilities across methods. \textbf{R}, \textbf{I}, and \textbf{M} indicate whether removal, insertion, and movement are supported within a single denoising step. \textbf{Img. Data} indicates that the model was trained on an image dataset.}
\label{tab:method_comparison}
\end{table}

\begin{figure*}[t!]
\centering
\includegraphics[width=0.9\linewidth]{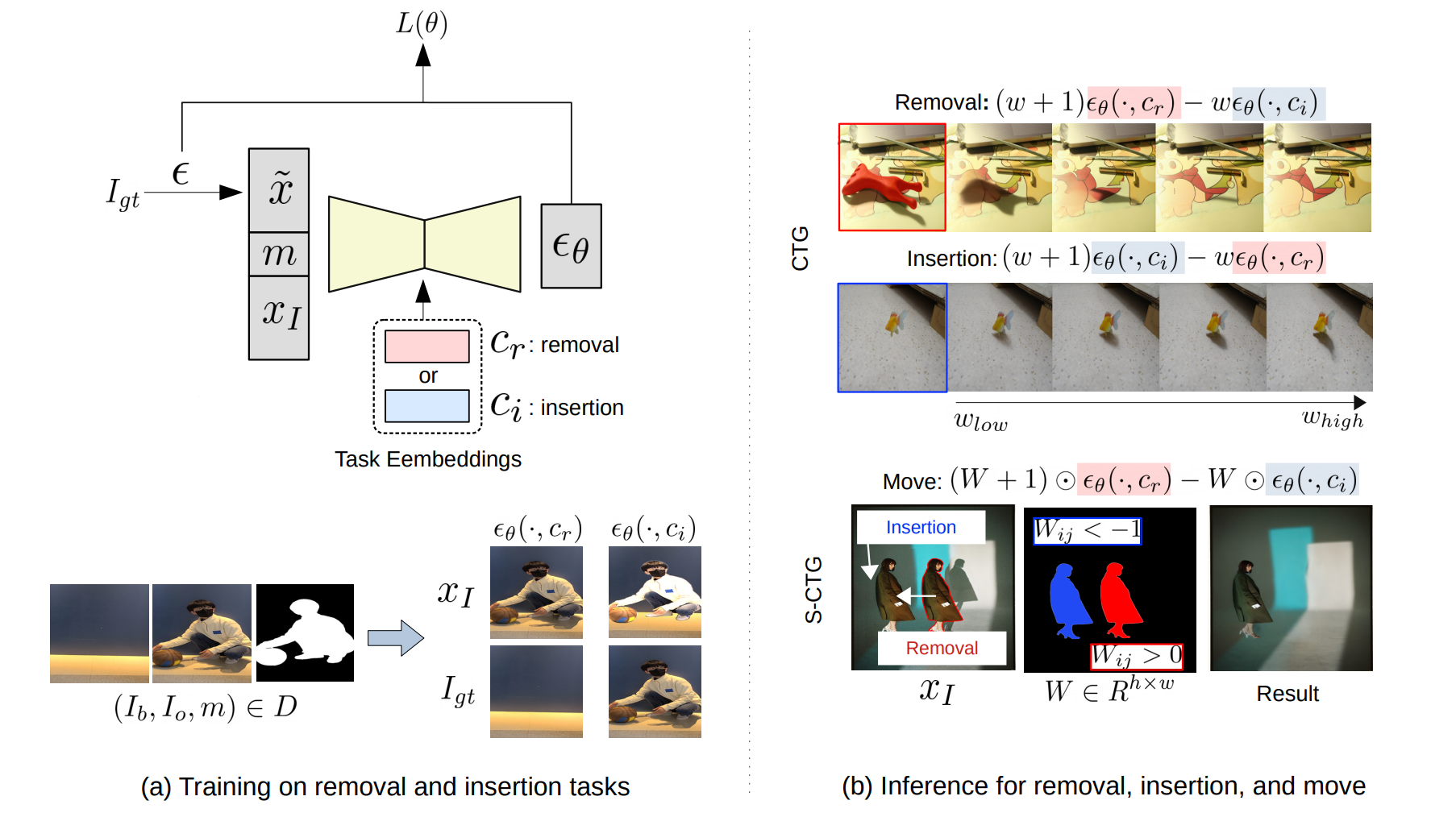}
\caption{
Using counterfactual $D$ and constant embeddings $\condremoval$ and $\condinsertion$, we train the diffusion network $\generator{}$ on two contrasting tasks: removal and insertion. Based on $\generator{}(\cdot, \condremoval)$ and $\generator{}(\cdot, \condinsertion)$, we propose cross-task guidance (CTG) which leverages classifier free guidance to better perform each task by using the other embeddings as the negative guidance. By proposing Spatial-CTG (S-CTG), we perform the two tasks in spatially different regions at a single denoising step, achieving the object movement task without additional training. 
}
\label{fig-train-inference}
   
\end{figure*}

\section{Method}
\subsection{Formulation}
\subsubsection{Task Definitions}
We build \algoname{} by developing a single diffusion model $\generator{}$ capable of performing the following three tasks:
\begin{itemize}
    \item \textbf{Removal}: Given an image $\condImage$ and a object-tight mask $\mask$,  $\generator{}$ removes the objects within the masked region along with their visual effects, such as shadows and reflections. Additionally, it offers user-controllable options to enhance the quality of removal.
    \item \textbf{Insertion}: 
    Given a copy-and-paste image $\condImage$ and a mask $\mask$ indicating the pasted region,  $\generator{}$ harmonizes the area by adapting local lighting and synthesizing plausible visual effects.
    We focus on a strict identity-preserving scenario, where $\mask$ corresponds exactly to the pasted pixels without any deformation.
    Users are also provided with control over the intensity and extent of visual effects.
    \item \textbf{Movement}: Given an image, we assume a set of pixels $\pixelRegionRemoval$ representing the object to be moved and a set $\pixelRegionInsertion$ indicating the insertion region. Their shapes are constrained to be identical, without deformation. The input image $\condImage$ contains the object already pasted at $\pixelRegionInsertion$, and the mask $\mask$ has non-zero values only over $\pixelRegionRemoval$ and $\pixelRegionInsertion$. The model $\generator{}$ removes the object and its associated visual effects from $\pixelRegionRemoval$, while harmonizing the pasted region and synthesizing plausible effects at $\pixelRegionInsertion$. Notably, the model performs both removal and insertion within a single denoising step, without separate processing phases. 
\end{itemize}
\subsubsection{Model}
In a similar manner to recent works~\cite{li2025roremtrainingrobustobject, zhao2025objectclearcompleteobjectremoval, objectdrop}, \algoname{} is built by fine-tuning the SDXL-inpainting model~\cite{podell2023sdxl}.
As shown in \figurename~\ref{fig-train-inference}-(a), the generator is defined as $\generator(x, t, \condImage, \mask, c)$, where $x$ is the noisy image, $\mask$ is a binary mask, $\condImage$ is the conditional image, $c$ is the prompt embedding, and $t$ denotes the denoising step. 
To simplify notation, we denote identical notation for both pixel- and latent-space representations of images.

\subsubsection{Dataset}
For fine-tuning, \algoname{} uses only the RORD dataset~\cite{sagong2022rord}, which provides a set of triplets $\data = \{(\imageBackground, \imageObject, \mask)\}$. Here, $\imageBackground$ refers to an image without the target object, while $\imageObject$ includes the object (see the bottom of Figure~\ref{fig-train-inference}-(a)). Since the original mask provided by the dataset is defined by coarse polygon shapes rather than object-tight boundaries, we compute a refined mask $m$ by extracting segmentation pixel groups contained within the polygon to ensure robust model performance.

\subsection{Training \algoname{}}
This section describes the training procedure for $\generator{}$ on object removal and insertion tasks. To eliminate the reliance on manually crafted text prompts, $\generator{}$ is trained using two fixed embeddings: $\condremoval$ for removal and $\condinsertion$ for insertion. 
The model is trained to reconstruct the ground-truth image $\image_{gt}$ given a noisy input $\tilde{x}$ and a conditional image $\condImage$.
We adopt the standard denoising diffusion loss:
\begin{equation}    
\mathcal{L}(\theta) =
{E}_{t \sim [0, T],\; \epsilon \sim \mathcal{N}(0, I)}
\left[
    \sum_{}
    \left\| 
        \generator{(\tilde{x},t,\condImage, \mask, c)} - \epsilon
    \right\|^2
\right]
\label{eqn-ssl}
\end{equation}
where $\tilde{x} = \alpha_t \image_{gt} + \sigma_t \epsilon$.

\subsubsection{Removal Task}
For the removal task, i.e., $\generator{(\cdot, \condremoval)}$, we set $\condImage = \imageObject$ and $\image_{gt} = \imageBackground$ in (\ref{eqn-ssl}) where $(\imageBackground, \imageObject, \mask) \in \data$.

\subsubsection{Insertion Task}
\begin{figure}[t]
\centering
\includegraphics[width=0.99\linewidth]{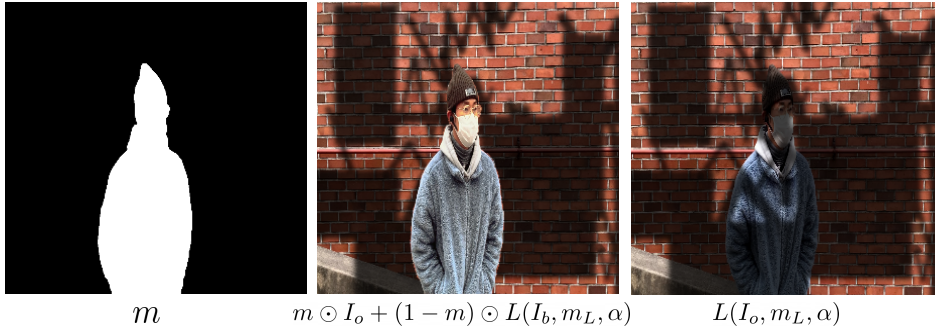}
\caption{
Shadow augmentation for enhanced  local light consistency of insertion task in Equation~(\ref{eqn: light-adjust}), with $\alpha<1$ in this example. $\condImage$ and $\image_{gt}$ are shown sequentially after the target mask $m$.
}
\label{fig-augmentation}   
\end{figure}

To train the model for the insertion task, i.e., $\generator{(\cdot, \condinsertion)}$, 
we provide $\condImage$ in which the object is copy-pasted into the background image with degraded harmonization.
Formally, we set the conditional image as:
\[
\condImage = \mask \odot \imageObjectAugmented + (1 - \mask) \odot \imageBackground,
\]
where $\odot$ denotes the element-wise multiplication and $C$ denotes a degrading operation, randomly chosen between Correlated Color Transform~\cite{cct} and intensity scaling.
To further reflect local light around the inserted region, 
we apply shadow augmentation to $\condImage$ by adjusting the intensity of a partial region of an image. Specifically,
for an image $\image$, we define the following shadowing operation:
\begin{equation}
L(\image, \mask_{L}, \alpha) = \mask_{L} \odot (\alpha \image) + (1-\mask_{L}) \odot \image,    
\label{eqn: light-adjust}
\end{equation}    
$\mask_{L}$ is a randomly shaped mask with Gaussian blur and $\alpha$ is an intensity scale.
Using the equation (\ref{eqn: light-adjust}) and triplet $(\imageBackground, \imageObject, \mask)$, we construct augmented training data for the insertion task as follows: $\condImage=\mask \odot \imageObjectAugmented  + (1-\mask) \odot L(\imageBackground, \mask_{L}, \alpha)$, $\image_{gt}= L(\imageObject, \mask_{L},  \alpha)$. 
\figurename~\ref{fig-augmentation} illustrates an example of the shadowing operation. 

\subsection{Inference with Controllability}
Building on $\generator{(\cdot, \condremoval)}$ and $\generator{(\cdot, \condinsertion)}$, we propose \textbf{Cross-Task Guidance (CTG)} to enhance the performance of both removal and insertion tasks through improved controllability. Furthermore, we introduce a spatially extended variant, referred to as \textbf{Spatial CTG (S-CTG)}, which enables removal and insertion to be performed on different regions within a single denoising step.

\subsubsection{Cross-task Guidance}
Relying solely on the conditional outputs $\generator{(\cdot, \condremoval)}$ and $\generator{(\cdot, \condinsertion)}$ may be insufficient to guarantee optimal performance or provide the controllability often needed to reflect user preferences.
Classifier-Free Guidance (CFG)\cite{cfg} and its variants, which leverage weaker model outputs\cite{pag, sg, autoguidance}, have demonstrated notable improvements in conditional generation by steering outputs away from unconditional or negatively conditioned references. 

In this context, we observe an inherent duality between removal and insertion prompts: the former removes object presence and its visual effects, while the latter synthesizes them. Building on this insight, we introduce \textbf{Cross-Task Guidance (CTG)} as follows:

\begin{align}
  \text{(Object removal)}   &\quad (1+\guidanceScale)\generatorRemoval{} - \guidanceScale \generatorInsertion{} , \label{eqn:removal}\\
  \text{(Object insertion)} &\quad (1 + \guidanceScale)\generatorInsertion{} - \guidanceScale \generatorRemoval{}, \label{eqn:insertion}
\end{align}
where $\guidanceScale > 0$ is a guidance scale, $\generatorRemoval{} = \generator{(\cdot, \condremoval)}$, and $\generatorInsertion{} = \generator{(\cdot, \condinsertion)}$.
Each task-specific output is guided by subtracting the output of the opposite task, effectively using it as negative guidance. 
As illustrated at the top of Figure~\ref{fig-train-inference}-(b), varying $\guidanceScale$ enables visual effect control.

\subsubsection{Spatial Cross-task Guidance for Movement Task}
A straightforward approach is to separate the movement task into two stages:
object removal followed by object insertion. 
Let $N_{{r}}$ and $N_{{i}}$ denote the number of denoising steps required to ensure visual quality in the removal and insertion phases, respectively. This sequential pipeline incurs a total of $N_{{r}} + N_{{i}}$ inference steps to maintain fidelity across both stages.
In contrast, our goal is to achieve object movement within $\max(N_{{r}}, N_{{i}})$ steps by jointly leveraging the removal and insertion embeddings, $\condremoval$ and $\condinsertion$, in a single denoising process. 
To this end, we extend CTG to operate in a region-specific manner.
Inspired by S-CFG~\cite{scfg}, which spatially adjusts the guidance scale of CFG based on semantic segments, we generalize the scalar guidance weight to a matrix form $W \in \mathbb{R}^{h \times w}$.

Specifically, CTG in Equation~(\ref{eqn:removal}) is applied to $\pixelRegionRemoval$ and CTG in Equation~(\ref{eqn:removal}) is applied to $\pixelRegionInsertion$.
As visualized in Figure~\ref{fig-train-inference}-(b),
S-CTG  assigns opposite signs to entries in $W$ to apply complementary CTGs to $\pixelRegionRemoval$ and $\pixelRegionInsertion$, thereby enabling efficient object movement within a single denoising step.
Considering the matrix-scaled guidance:
\begin{equation}
(1+W) \odot \generatorRemoval{} - W  \odot \generatorInsertion{},    
\label{eqn:s-ctg}
\end{equation}
we determine each element $W_{ij}$ of the matrix $W$ based on the pixel's membership in $\pixelRegionRemoval$ or $\pixelRegionInsertion$, as well as the current denoising step $t$:
\begin{equation}
    W_{ij} =
\begin{cases}
w_r & \text{if } (i,j) \in \mathcal{P}_r \\
w_i & \text{if } (i,j) \in \mathcal{P}_i \\
\begin{cases}
w_r & \text{if } t > t_{s} \\
w_i & \text{if } t < t_{s}
\end{cases} & \text{if }(i,j) \in \mathcal{P} \setminus (\mathcal{P}_r \cup \mathcal{P}_i)
\end{cases},
\end{equation}
where $\mathcal{P}$ denotes the set of all pixels, and $w_r > 0$, $w_i < -1$ are the guidance weights for removal and insertion, respectively. The switching step $t_s$ adjusts the \textit{gray area}—pixels belonging to neither $\mathcal{P}_r$ nor $\mathcal{P}_i$—to prioritize removal during the early denoising steps and insertion during the later steps. 

\section{Experiments}
\subsection{Implementations}
\subsubsection{Training Details}
We fine-tune the model on the RORD dataset with a 2:1 ratio of removal to insertion samples, a learning rate of 1e-5, and an effective batch size of 512 using gradient accumulation. Training is conducted with the Adam optimizer on 8 NVIDIA H100 GPUs.
\subsubsection{Evaluation  Details}
We conduct experiments using a test set with available ground-truth or reference images, and accordingly employ reference-based evaluation metrics such as LPIPS, PSNR, CLIP \cite{clipscore} image score, DINO \cite{dinoscore} image score, and FID.
For all evaluations, we report the average 
scores across 10 different random seeds. 
For in-mask only editing models, we apply additional mask dilation to ensure fair evaluation. For models trained with counterfactual dataset, we fix the dilation size to 10 pixels.
We report all parameter tuning details and additional visual results in the Supplementary sections.

\subsection{Object Removal}

We conduct benchmark experiments for the removal task to evaluate two key aspects: \textbf{R1)} object removal performance of \algoname{} and \textbf{R2)} the controllability provided by adjusting the CTG scale.
For comparison, we employ five baseline methods, categorized as follows: two in-mask editing methods: PowerPaint and ClipAway~\cite{powerpaint, ekin2024clipawayharmonizingfocusedembeddings}; one inversion-based method: AttentiveEraser~\cite{sun2025attentiveeraserunleashingdiffusion}; and two counterfactual-based methods: ROREM and OmniEraser~\cite{li2025roremtrainingrobustobject, wei2025omnieraserremoveobjectseffects}.


We use \textbf{RemovalBench}~\cite{li2025roremtrainingrobustobject} as a benchmark for object removal.
To address more challenging real-world scenarios, we introduce \textbf{BenchHard}, which consists of 100 images—50 drawn from the RORD validation set and 50 captured by professional photographers. Compared to RemovalBench, BenchHard presents more challenging scenarios characterized by (1) larger masks, (2) object removal within repetitive patterns, and (3) complex shadows and reflections. We attached the 100 images  in the Supplementary Materials.

\begin{figure*}[h!]
\centering
\includegraphics[width=0.9\linewidth]{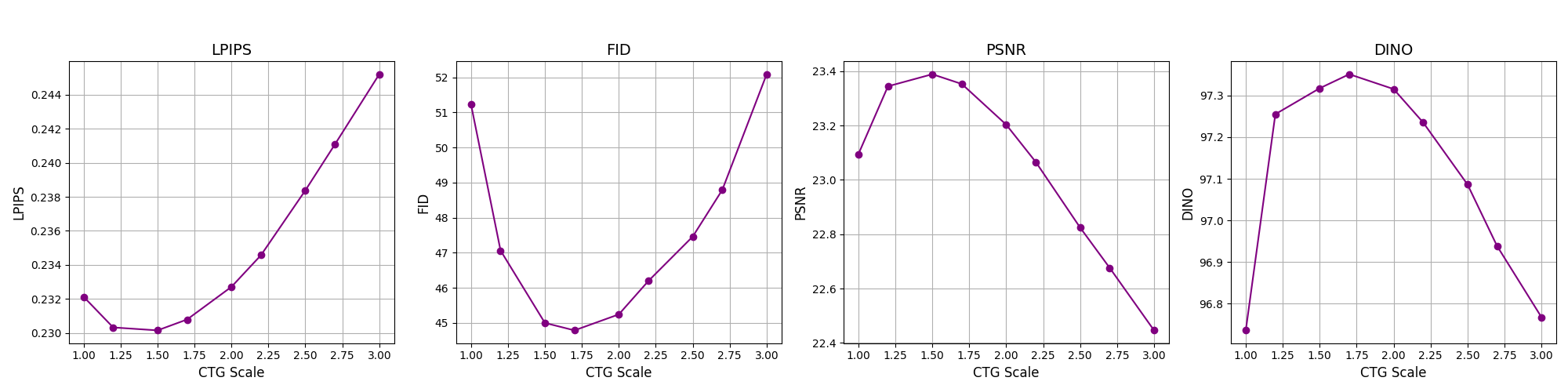}    
\caption{Object removal performance on BenchHard along with CTG scales.}
\label{fig:ctg-scale}
\end{figure*}

Figure~\ref{fig-removal} and Table~\ref{tab:removal-comparison} show qualitative and quantitative results, respectively, with the CTG scale $w$ to 1.5.
It can be observed that \textbf{(R1)} \algoname{} outperforms the baselines across most metrics in both benchmarks.
As shown in Figure~\ref{fig-removal}, CTG enables the effective removal of both the object and its associated visual effects.
To demonstrate \textbf{R2}, we plot the results across different CTG scales in Figure~\ref{fig:ctg-scale}.
We can clearly observe that the results vary with the CTG scale, with a value around 1.5 consistently yielding strong performance across all metrics.
These results demonstrate that CTG offers reliable task-specific guidance and enables users to flexibly control the output.


\begin{table*}[t]
\centering
\caption{Comparison of object removal performance across two benchmarks.}
\label{tab:removal-comparison}
\begin{tabular}{l|cccc|cccc}
\toprule
\multirow{2}{*}{Algorithm} & \multicolumn{4}{c|}{\textbf{RemovalBench} \cite{li2025roremtrainingrobustobject}} & \multicolumn{4}{c}{\textbf{BenchHard}} \\
 & LPIPS ↓ & PSNR ↑ & CLIP ↑ & DINO ↑ & LPIPS ↓ & PSNR ↑ & CLIP ↑ & DINO ↑ \\
\midrule
PowerPaint & 0.346 & 22.208 & 96.384 & 90.791 & 0.326 & 19.505 & 96.369 & 90.954 \\
CLIPAway & 0.396 & 22.207 & 96.401 & 88.934 & 0.378 & 19.235 & 96.124 & 90.653 \\
OmniEraser & 0.377 & 22.648 & 97.528 & 94.156 & 0.366 & 21.146 & 97.255 & 93.906 \\
Attentive Eraser & \textbf{0.256} & {24.515} & 98.072 & 94.391 & \underline{0.235} & 22.446 & 97.937 & 94.416 \\
RORem & \underline{0.258} & 24.262 & 98.130 & 94.223 & 0.243 & 22.417 & 98.010 & 94.613 \\
\rowcolor{green!10} CrimEdit (w/o CTG) & 0.272 & \textbf{24.638} & \underline{98.787} & \underline{96.674} & 0.236 & \underline{23.194} & \underline{98.403} & \underline{96.744} \\
\rowcolor{green!10} CrimEdit (w/ CTG) & {0.271} & \underline{24.630} & \textbf{98.794} & \textbf{96.717} & \textbf{0.234} & \textbf{23.249} & \textbf{98.523} & \textbf{97.176} \\
\bottomrule
\end{tabular}
\end{table*}

\begin{figure}[t!]
\centering
\includegraphics[width=0.98\linewidth]{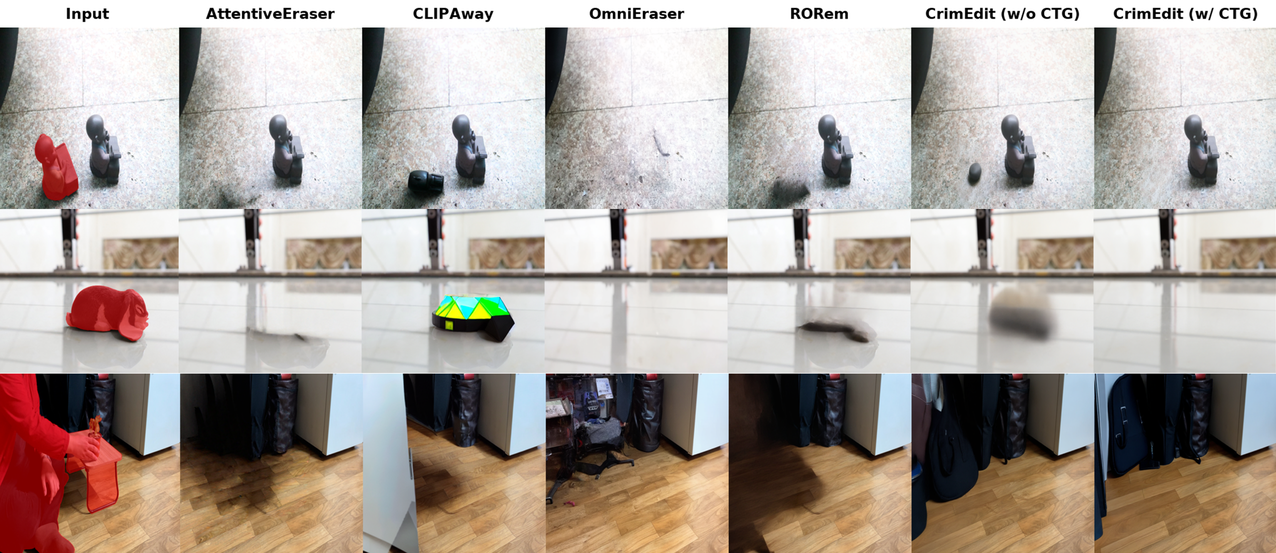}
\caption{
Counterfactual object removal results on BenchHard, which includes challenging scenarios such as repetitive objects (first row), strong reflections (second row), and large mask regions (third row), are shown.
} 
\label{fig-removal}   
\end{figure}

\begin{figure}[t!]
\centering
\includegraphics[width=0.95\linewidth]{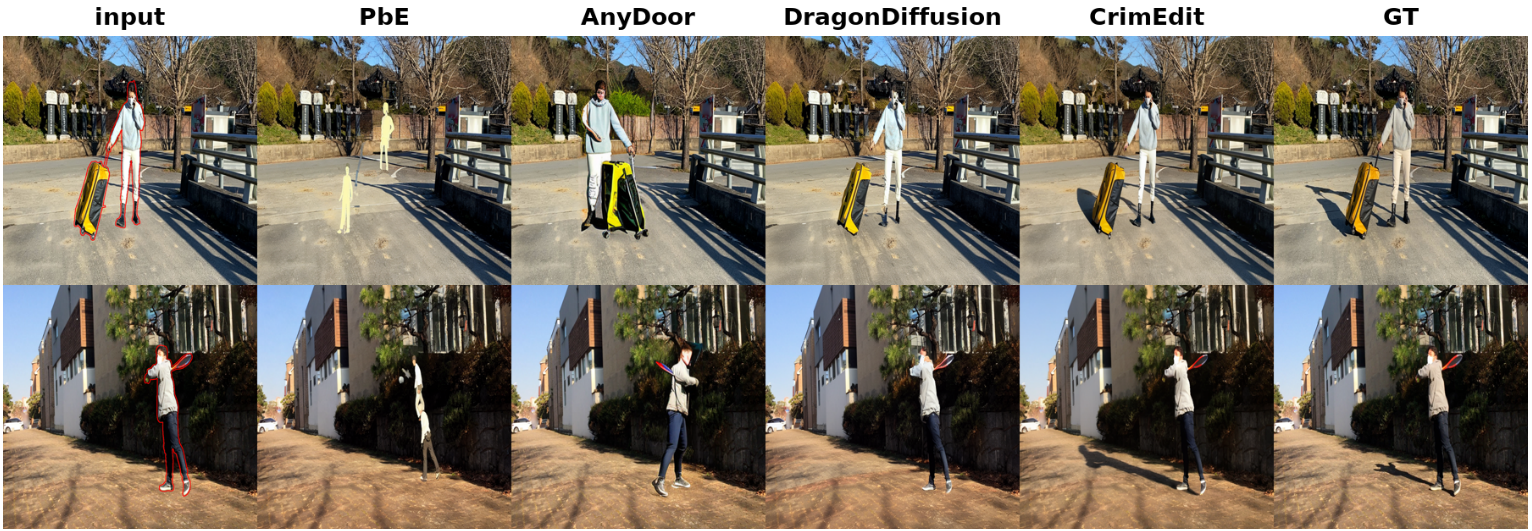}  
\includegraphics[width=0.95\linewidth]{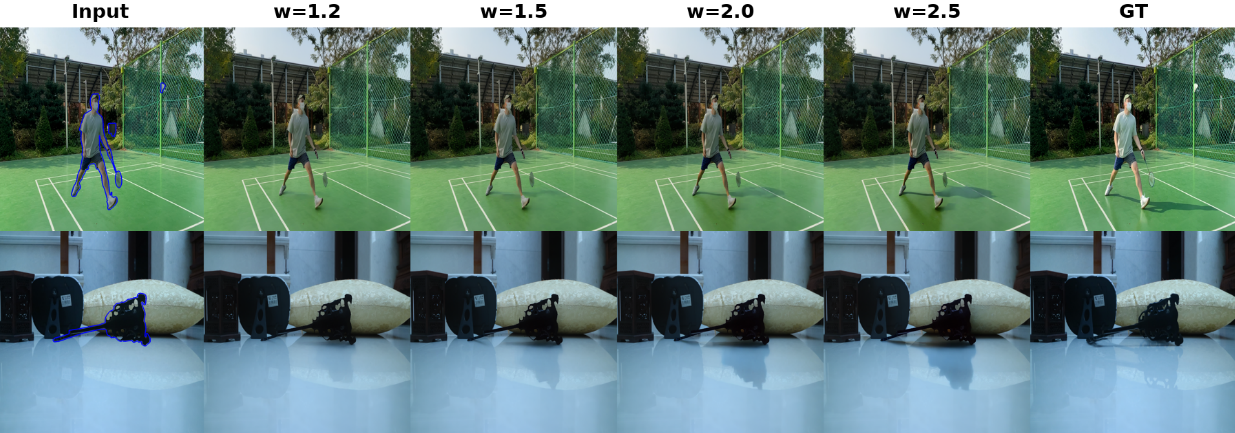}    

\caption{(Top) Object insertion comparison result. (Bottom) Controllability tests on object insertion task.}
\label{fig:ctg-scale-hmz}
\end{figure}

\begin{figure}[t]
\centering
\includegraphics[width=0.95\linewidth]{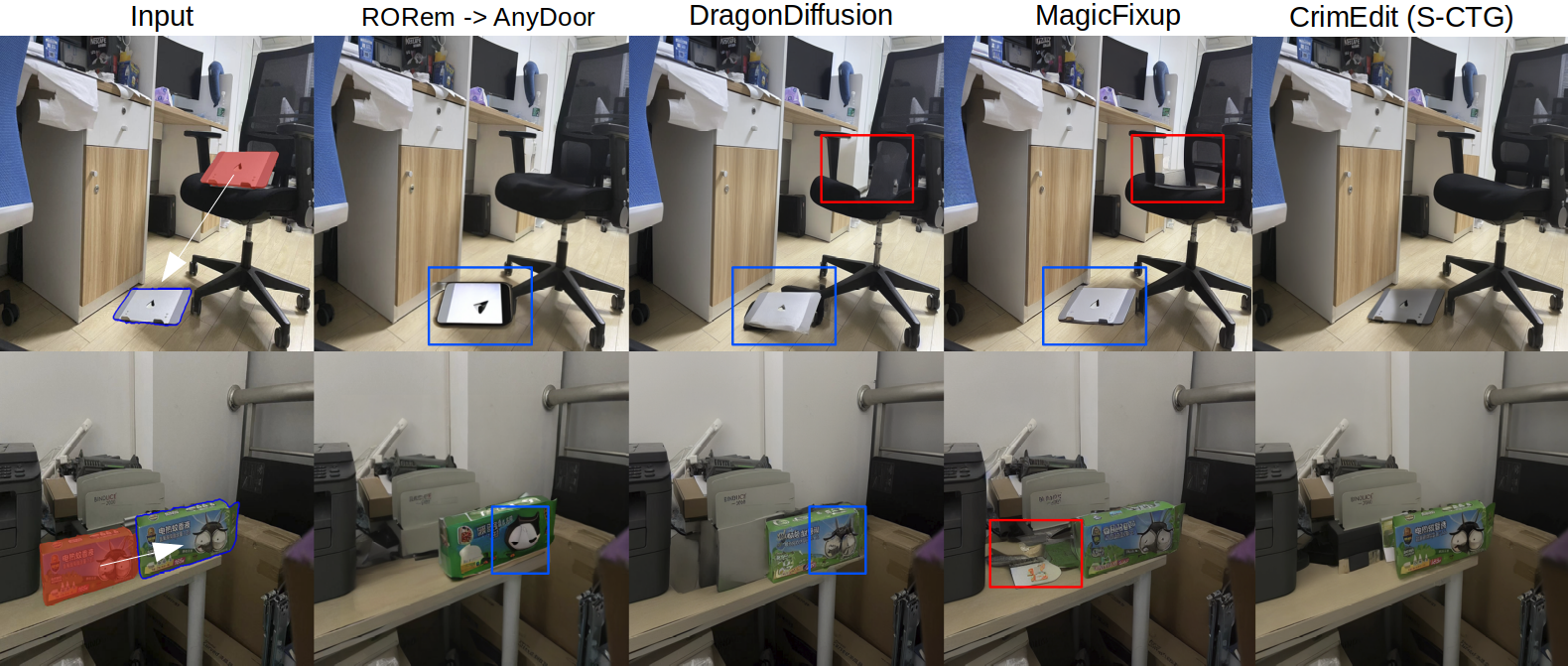}
\includegraphics[width=0.95\linewidth]{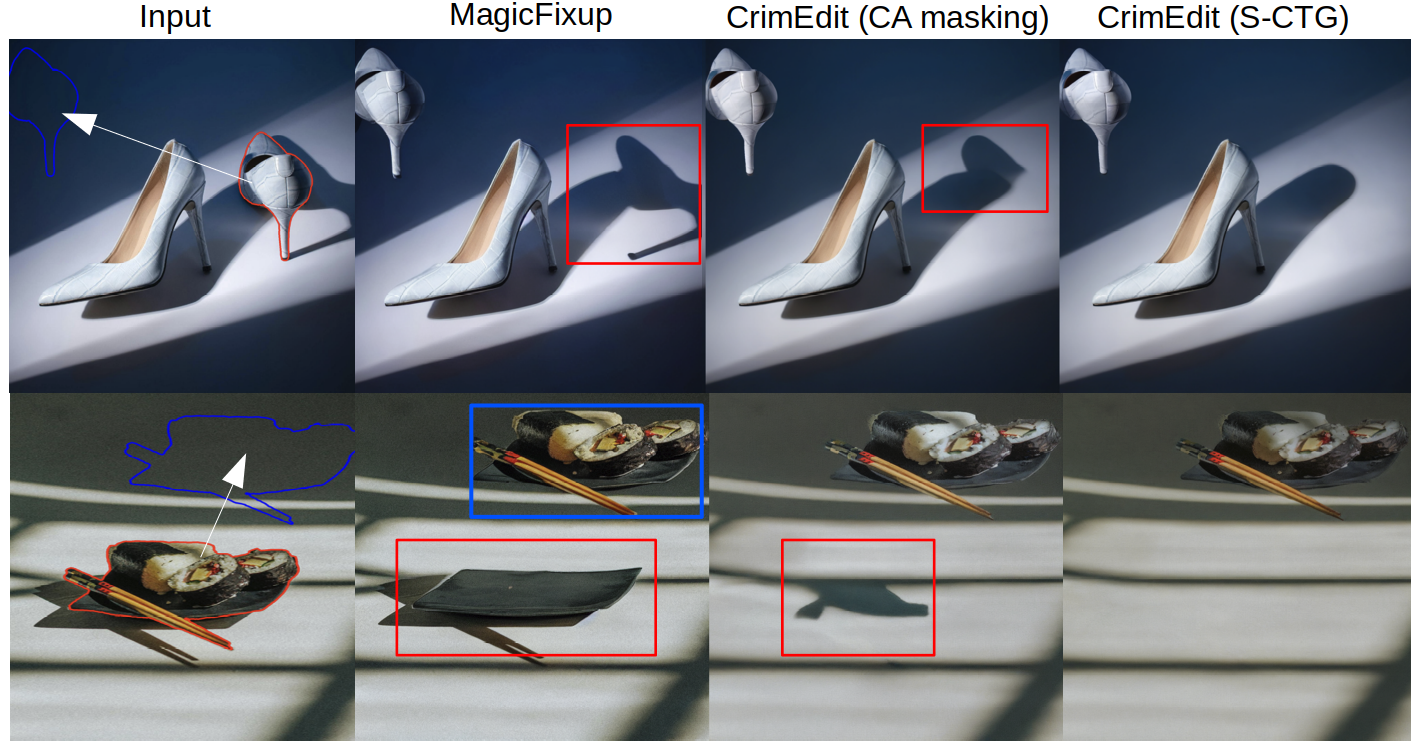}
\caption{
Red boxes indicate unrealistic removal artifacts, while blue boxes denote unrealistic harmonization or effects generation.
(Top) Comparison of object movement results on the ReS dataset. 
(Bottom) Qualitative comparison between CA masking and S-CTG.
}
\label{fig: res-move-comparison}   
\end{figure}

\subsection{Object Insertion}

To evaluate object insertion performance, we use the RORD validation set, assessing both color harmonization and visual effect generation.
To this end, we construct the conditional images ($\condImage$) by copy-pasting target objects from the ground-truth image ($\imageObject$) into the background image ($\imageBackground$) and applying a color transform to the pasted object.
We investigate two key aspects:
\textbf{I1}) whether $\generatorInsertion{}$ with CTG achieves performance comparable to existing methods, and \textbf{I2}) whether the CTG scale effectively controls the strength of visual effects.

As shown in Table~\ref{tab: Object insertion comparative result} and Figure~\ref{fig:ctg-scale-hmz}, \algoname{} synthesizes more realistic visual effects while better preserving object identity compared to existing methods, thereby validating \textbf{I1}.
Figure~\ref{fig:ctg-scale-hmz} also demonstrates \textbf{I2}, where
larger guidance scales produce correspondingly larger shadows and reflections. 
In addition, we measure how the shadow area changes with different CTG scale values, using LISA for shadow detection.
In Table~\ref{tab: Shadow Generation Results}, 
PSNR and LPIPS are computed between the output and the original input $\condImage$, restricted to the unmasked region. A larger difference indicates a greater degree of change due to visual effect synthesis.
We can observe that larger shadow regions are generated under increased CTG scaling. 
\begin{table}[h]
\centering
\caption{Object insertion comparative result}
\label{tab:hmz}
\begin{tabular}{lllll}
\toprule
 & LPIPS ↓ & PSNR ↑ & CLIP ↑ & DINO ↑ \\
\midrule
PbE & 0.418 & 17.401 & 93.238 & 85.602 \\
AnyDoor & 0.406 & 17.252 & 94.536 & 89.556 \\
FreeComp. & 0.390 & 19.486 & 93.300 & 89.761 \\
Drag.Diff. & \underline{0.258} & \underline{21.039} & \underline{96.984} & \underline{94.012} \\
\rowcolor{green!10} CrimEdit & \textbf{0.228} & \textbf{22.174} & \textbf{98.055} & \textbf{97.452} \\
\bottomrule
\end{tabular}

\label{tab: Object insertion comparative result}
\end{table}

\begin{table}[h]
\setlength{\tabcolsep}{4pt}
\centering
\caption{Shadow Generation Results}
\begin{tabular}{lcccc}
\toprule
\makecell{CTG \\ scale} & \makecell{Avg. Shad. \\ Pixel Ratio (\%)}     & \makecell{Ratio of \\ Shad. Det.  }    & \makecell{PSNR }     & \makecell{LPIPS} \\ \midrule

1.2 & 0.336 & 0.27 & 31.57 & 0.1741 \\
1.7 & 0.532 & 0.33 & 31.00 & 0.1848 \\
2.0   & 0.625 & 0.41 & 30.61 & 0.1941 \\
2.5 & 0.776 & 0.42 & 30.09 & 0.2163 \\ \bottomrule

\end{tabular}
\label{tab: Shadow Generation Results}

\end{table}

\subsection{Object Movement}
We evaluate object movement performance using the ReS dataset~\cite{seele}.
This evaluation focuses on three aspects: 
\textbf{M1}) whether \algoname{} with S-CTG achieves comparable performance to prior methods in counterfactual object repositioning, 
\textbf{M2}) whether it outperforms the cross-attention (CA) masking strategy from FunEditor~\cite{funeditor},
and \textbf{M3}) whether the light adjustment technique in Eq.~(\ref{eqn: light-adjust}) improves local harmonization.
Regarding \textbf{M1}, we compare \algoname{} with two open-sourced methods \cite{dragondiffusion, alzayer2024magicfixup}.
Additionally,  \algoname{} is compared with the result obtained by using separate models for object removal \cite{li2025roremtrainingrobustobject} and insertion \cite{anydoor}. 
The result is shown Table~\ref{tab: Object Movement} and Figure~\ref{fig: res-move-comparison}. As shown in Table~\ref{tab: Object Movement},
\algoname{} with S-CTG outperforms other methods with efficient computation times
(\textbf{M1}).
In terms of 
\textbf{M2}, Table \ref{tab: Object Movement} and Figure \ref{fig: res-move-comparison} also show the performance gain when using S-CTG rather than CA masking when the same model is used. 
\textbf{M3} is also confirmed by Table \ref{tab:movement} where S-CTG shows better performance than the model trained for the same step without light augmentation. 
More visual results can be found in supplementary material. 

 \begin{table}[h!]
 \setlength{\tabcolsep}{4pt}
\centering
\caption{Object Movement}
\label{tab:movement}
\begin{tabular}{lllll}
\toprule
 & LPIPS ↓ & PSNR ↑ & DINO ↑ & \makecell{T[s] ↓ } \\
\midrule
Anydoor & 0.374 & 18.402 & 92.501 & 13.55 \\
Drag.Diff. & 0.304 & 18.294 & 93.404 & 16.20 \\
MagicFixup & 0.374 & \textbf{19.078} & 94.384 & \textbf{4.25} \\
\rowcolor{green!10} Ours (w/o Aug.) & 0.289 & 18.954  & 94.942 & 6.74\\
\rowcolor{green!10} Ours (CA) & \underline{0.291} & \underline{19.057} & \underline{95.026} & {\underline{6.65}} \\
\rowcolor{green!10} Ours (S-CTG) & \textbf{0.286} & 19.031 & \textbf{95.092} & 6.74 \\
\bottomrule
\end{tabular}
\label{tab: Object Movement}
\end{table}

\section{Conclusion}
This paper presented a unified diffusion model, \algoname{}, that addresses three image editing tasks—counterfactual object removal, insertion, and movement—by training on two task-specific embeddings, $\condremoval$ and $\condinsertion$, derived from a counterfactual dataset. Through the use of Cross-Task Guidance (CTG), \algoname{} demonstrates robust object removal and controllable insertion, particularly in handling visual effects such as shadows. Furthermore, the extension of CTG into its spatial variant, S-CTG, achieves superior performance in object repositioning compared to alternative approaches, including cross-attention masking and separate-stage pipelines

{
    \small
    \bibliographystyle{ieeenat_fullname}
    \bibliography{main}
}

\clearpage
\onecolumn
\begin{center}
    \Large \textbf{Supplementary Material}
\end{center}

\section{Additional Ablations}

\subsection{Light augmentation in Movement}
\begin{figure*}[h!]
    \centering
    \includegraphics[width=0.99\linewidth]{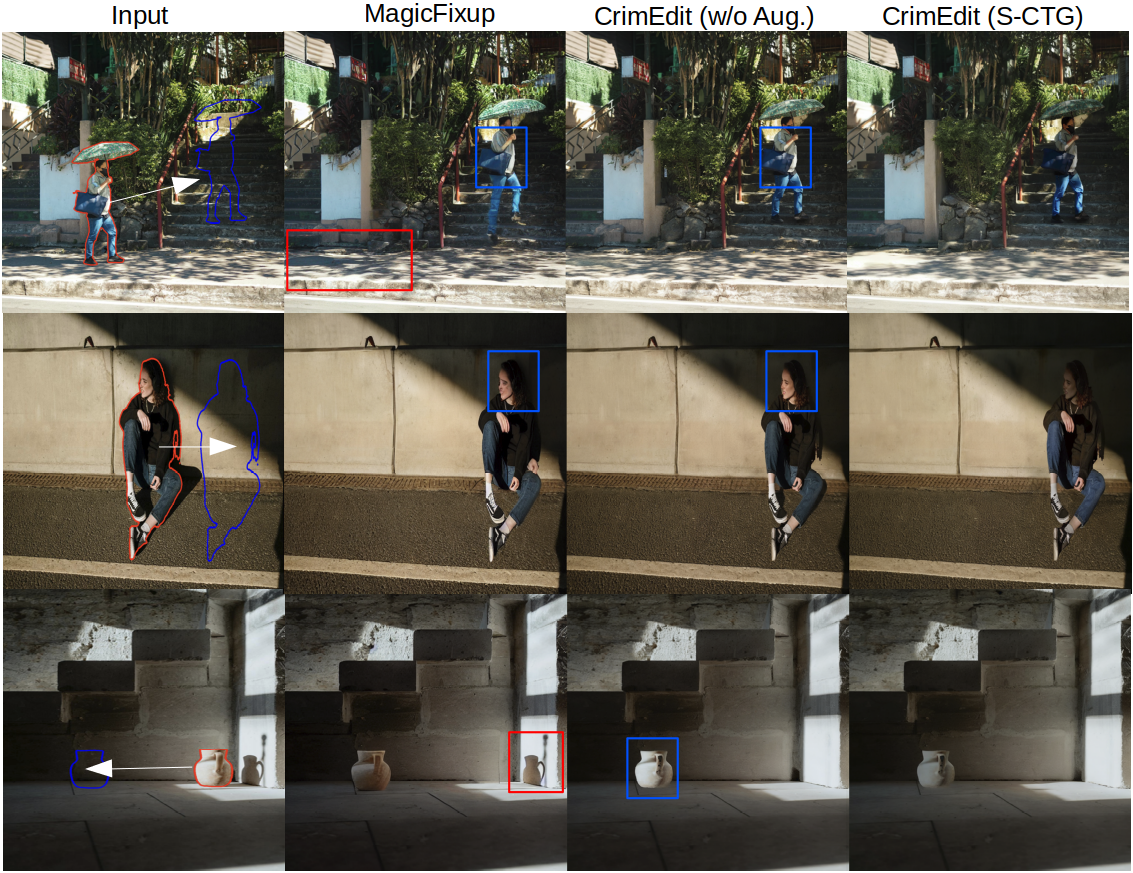}
    \caption{Comparison result for the light augmentation.}
    \label{fig:aug-comp}
\end{figure*}

As stated in equation (2), CrimEdit adopted optional lighting augmentation. Qualitative comparison is shown in Figure~\ref{fig:aug-comp}.



\subsection{Separate-phased Approach in Movement}
\begin{figure}[h!]
    \centering
    \includegraphics[width=0.9\linewidth]{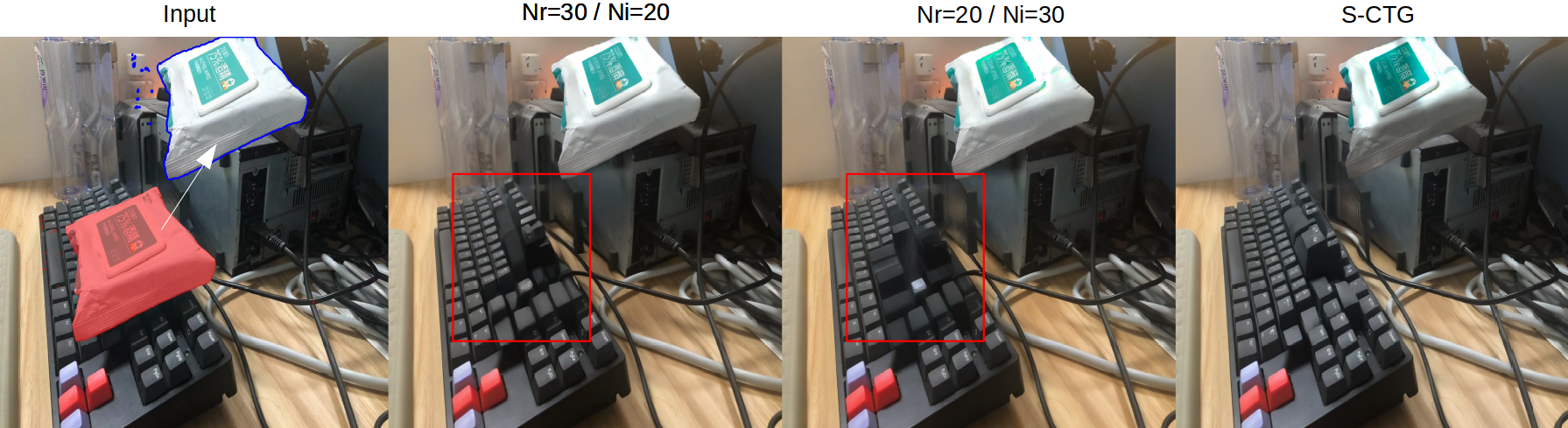}
    \caption{Visual comparison with separate-phased approach. The box regions show that small amount of denoising steps for removal can lead to blurry result.}
    \label{fig:sperate-stage-1}
\end{figure}

\begin{figure}[h!]
    \centering
    \includegraphics[width=0.9\linewidth]{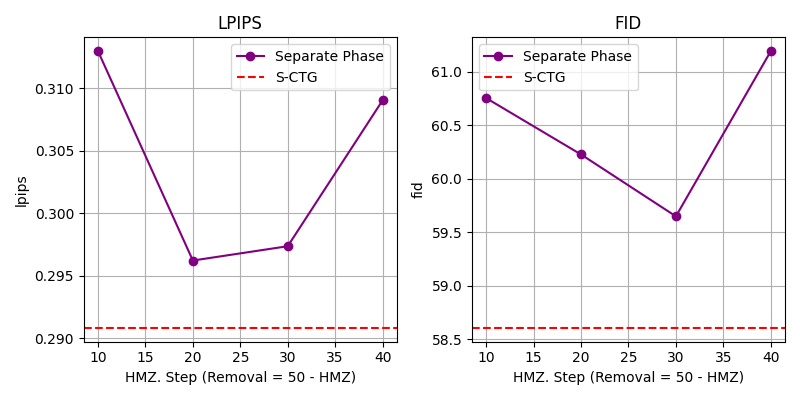}
    \caption{Comparison with separate-phased approach.}
    \label{fig:sperate-stage-2}
\end{figure}

Using ReS dataset, we compare S-CTG with the separate-phased approach consisting of object removal and insertion. 
The latter approach is given the total $N_r + N_i = 50$ denoising step where $N_r$ is the steps used for removal and $N_i$ for insertion. We used the same model and CTG scale $1.5$ was applied for each phase. Figure \ref{fig:sperate-stage-1} and \ref{fig:sperate-stage-2} shows that S-CTG yielded better results for the visual quality and perceptual metrics.

\section{Additional Qualitative Results}
\begin{figure}[h!]
    \centering
    \includegraphics[width=0.9\linewidth]{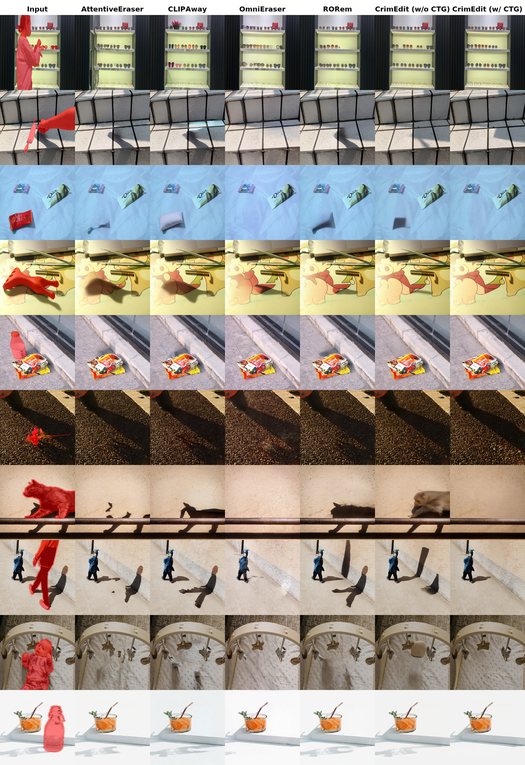}
    \caption{Qualitative comparison for the counterfactual removal for BenchHard dataset. The models are given the red-shaded mask only, without including the visual effects. }
    \label{fig:removal-comp}
\end{figure}

\begin{figure}[h!]
    \centering
    \includegraphics[width=0.9\linewidth]{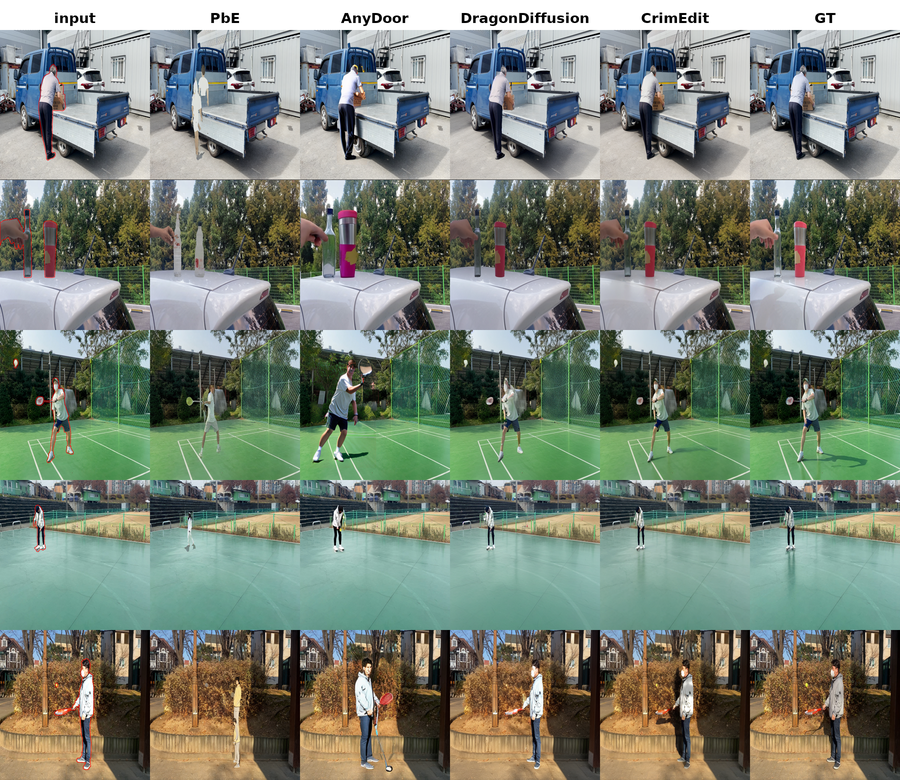}
    \caption{Qualitative comparison for the counterfactual insertion for RORD validation set.}
    \label{fig:hmz-comp}
\end{figure}

\begin{figure}[h!]
    \centering
    \includegraphics[width=0.9\linewidth]{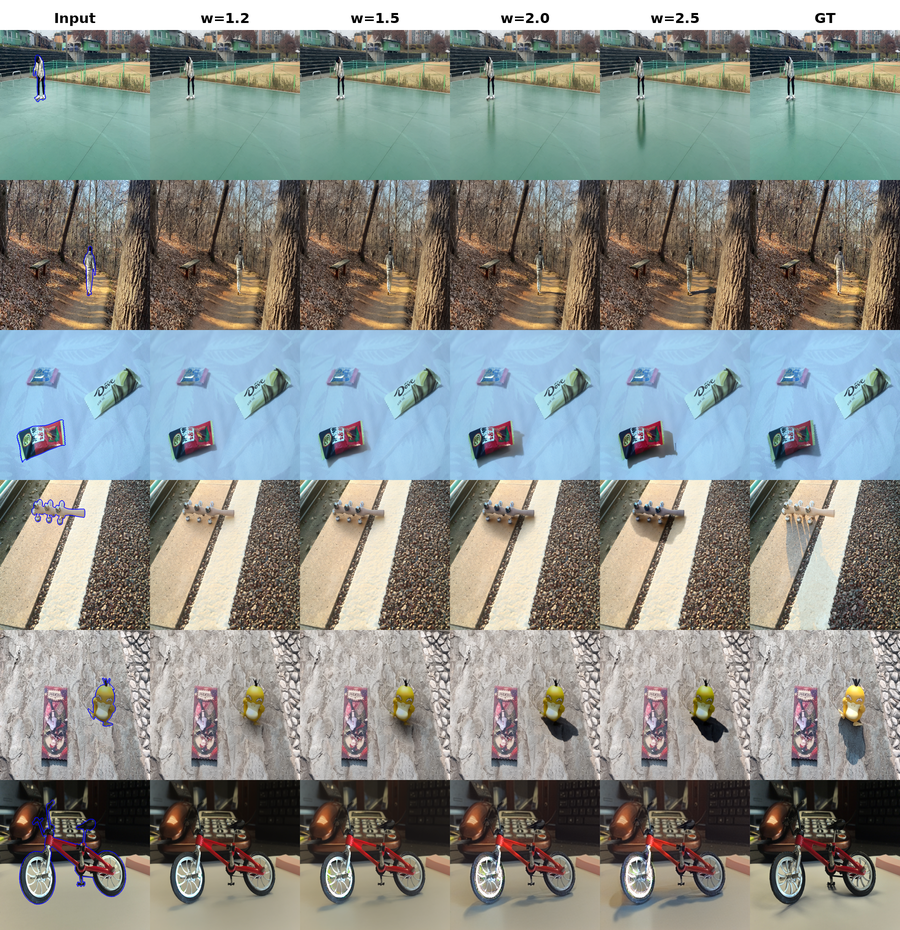}
    \caption{Controllability test for insertion task by changing the CTG weight.}
    \label{fig:hmz-control}
\end{figure}

In this section, we include more visual results to access the proposed methods. Figure \ref{fig:removal-comp} shows the comparative results showing the superior performance of the proposed CTG method. Figure \ref{fig:hmz-comp} and Figure \ref{fig:hmz-control} depict the comparative result and the effect of the CTG scale in the insertion task. As Figure \ref{fig:hmz-control} shows, a higher CTG scale produces larger shadow mask. 

\begin{figure}[h!]
    \centering
    \includegraphics[width=0.85\linewidth]{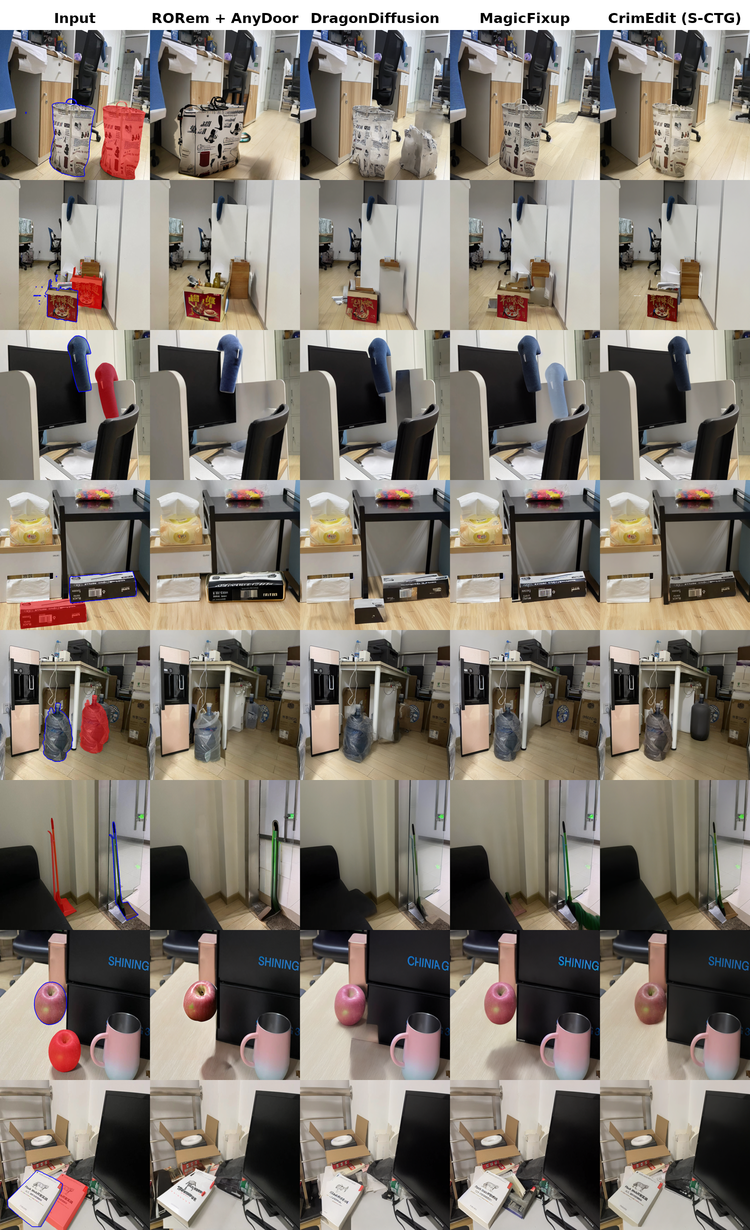}
    \caption{Object movement comparative results. }
    \label{fig:movement-control}
\end{figure}

\section{Evaluation Details}
\subsection{BenchHard Dataset}
\begin{figure*}
    \centering
    \includegraphics[width=0.99\linewidth]{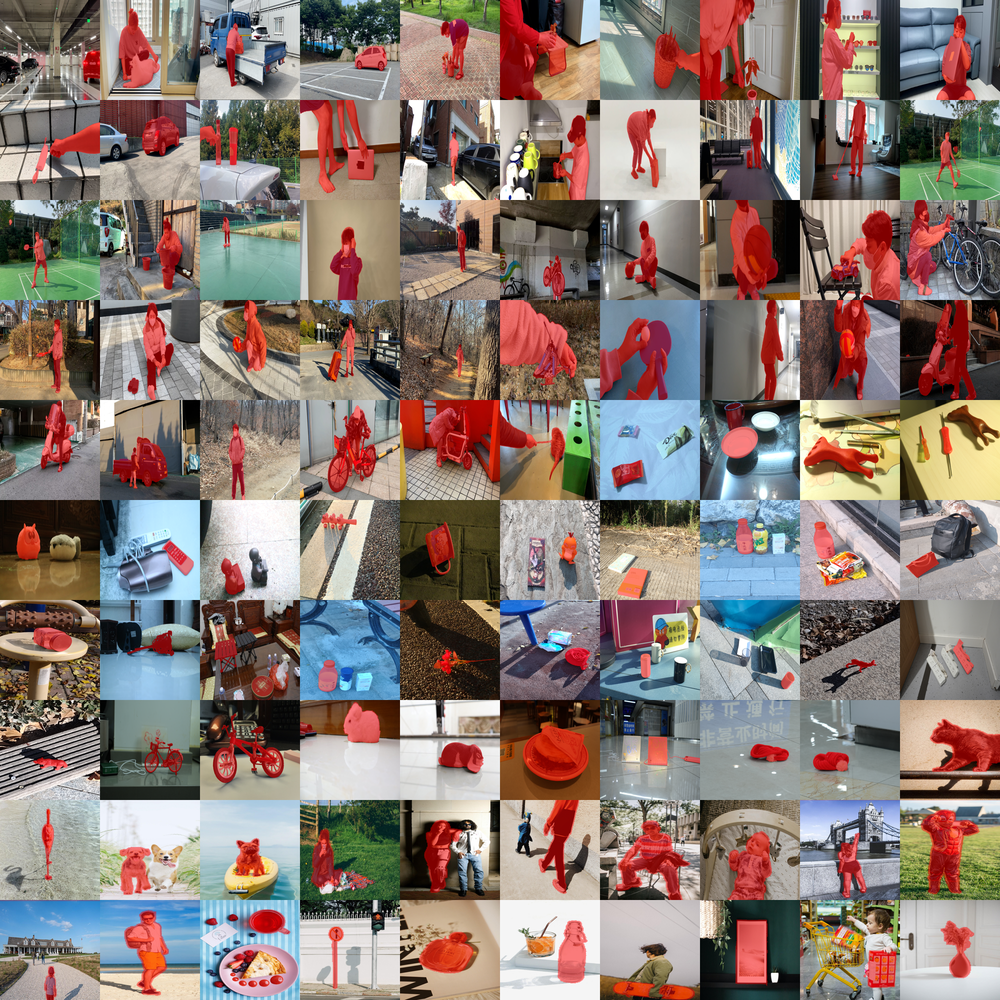}
    \caption{100 images of BenchHard. Red shade visualizes the given masks.}
    \label{fig:enter-label}
\end{figure*}

\begin{figure*}
    \centering
    \includegraphics[width=0.99\linewidth]{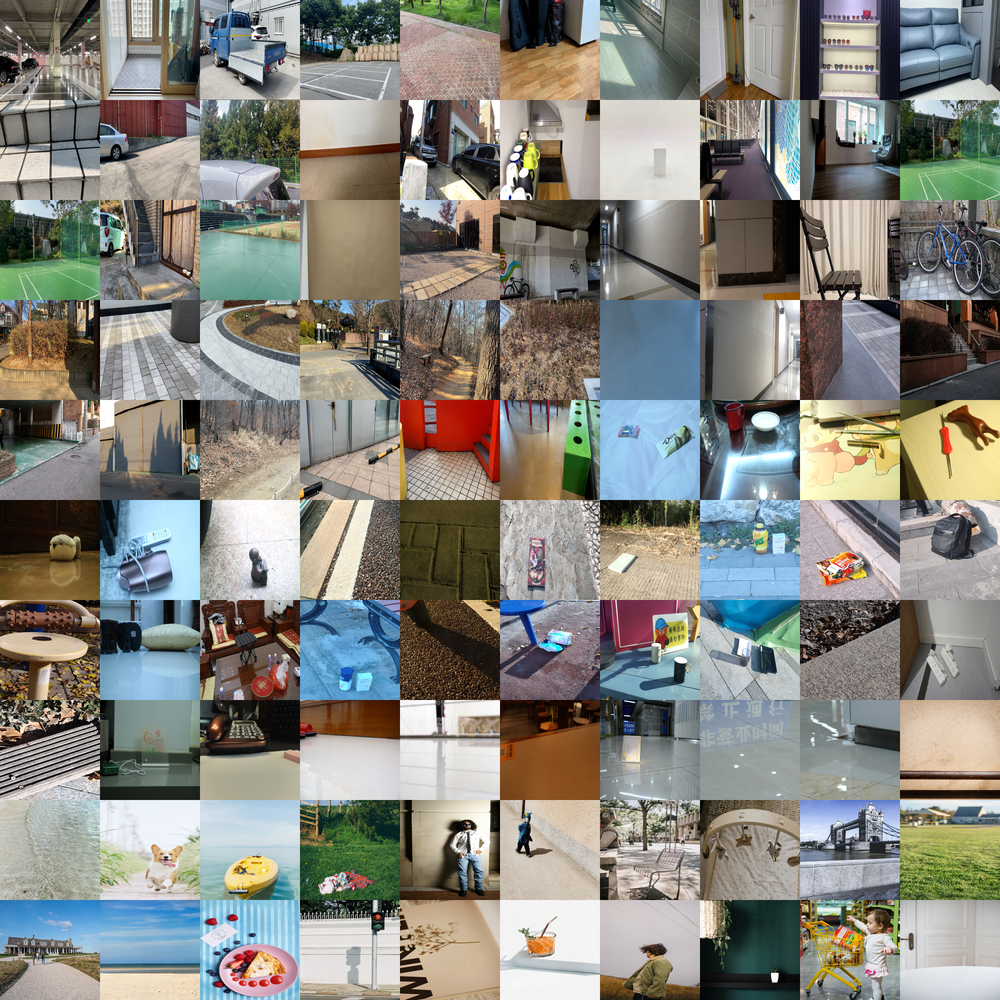}
    \caption{100 ground truth images of BenchHard.}
    \label{fig:enter-label-2}
\end{figure*}

\subsection{Parameter Tuning of Baselines}
We tuned the other arts by finding the parameter set that best rewards the sum of the metric set in a normalized scale. This section shows the turning history and their selected parameters, shown in a boldface. 

\begin{table*}[ht]
\centering
\caption{PowerPaint (Left: BenchHard / Right: RemovalBench)}
\setlength{\tabcolsep}{3pt} 
\begin{minipage}[t]{0.48\textwidth}
\centering
\begin{tabular}{ll|lllll}
\toprule
Dilation & Scale & LPIPS & FID & PSNR & CLIP & DINO \\
\midrule
10 & 10.5 & 0.3255 & 108.09 & 19.40 & 96.24 & 90.81 \\
10 & 13.5 & 0.3291 & 108.89 & 19.28 & 96.10 & 90.65 \\
\textbf{15} & \textbf{10.5} & \textbf{0.3264} & \textbf{105.52} & \textbf{19.50} & \textbf{96.37} & \textbf{90.95} \\
15 & 13.5 & 0.3300 & 107.75 & 19.35 & 96.21 & 90.76 \\
\bottomrule
\end{tabular}
\end{minipage}
\hfill
\begin{minipage}[t]{0.48\textwidth}
\centering
\begin{tabular}{ll|lllll}
\toprule
Dilation & Scale & LPIPS & FID & PSNR & CLIP & DINO \\
\midrule
10 & 10.5 & 0.3470 & 135.22 & 21.95 & 96.27 & 90.18 \\
10 & 13.5 & 0.3515 & 142.27 & 21.71 & 96.11 & 89.94 \\
\textbf{15} & \textbf{10.5} & \textbf{0.3462} & \textbf{123.78} & \textbf{22.21} & \textbf{96.38} & \textbf{90.79} \\
15 & 13.5 & 0.3511 & 135.00 & 21.92 & 96.14 & 90.53 \\
\bottomrule
\end{tabular}
\end{minipage}
\end{table*}

\begin{table*}[ht]
\centering
\caption{CLIPAway Results: (Left: BenchHard / Right: RemovalBench)}
\setlength{\tabcolsep}{3pt} 
\begin{minipage}[t]{0.48\textwidth}
\centering
\begin{tabular}{ll|lllll}
\toprule
Dilation & Scale & LPIPS & FID & PSNR & CLIP & DINO \\
\midrule
10 & 0.5 & 0.3751 & 112.34 & 19.30 & 96.11 & 89.84 \\
\textbf{10} & \textbf{1.0} & \textbf{0.3783} & \textbf{110.66} & \textbf{19.23} & \textbf{96.12} & \textbf{90.65} \\
10 & 1.5 & 0.3822 & 109.55 & 19.07 & 96.06 & 90.51 \\
15 & 0.5 & 0.3778 & 113.04 & 19.24 & 96.07 & 89.64 \\
15 & 1.0 & 0.3810 & 112.24 & 19.21 & 96.11 & 90.44 \\
15 & 1.5 & 0.3864 & 114.86 & 18.98 & 95.92 & 90.13 \\
\bottomrule
\end{tabular}
\end{minipage}
\hfill
\begin{minipage}[t]{0.48\textwidth}
\centering
\begin{tabular}{ll|lllll}
\toprule
Dilation & Scale & LPIPS & FID & PSNR & CLIP & DINO \\
\midrule
10 & 0.5 & 0.3957 & 132.51 & 22.12 & 96.29 & 88.40 \\
10 & 1.0 & 0.3980 & 130.70 & 22.05 & 96.32 & 89.12 \\
10 & 1.5 & 0.4012 & 133.89 & 21.76 & 96.09 & 89.13 \\
\textbf{15} & \textbf{0.5} & \textbf{0.3957} & \textbf{121.60} & \textbf{22.21} & \textbf{96.40} & \textbf{88.93} \\
15 & 1.0 & 0.3981 & 122.68 & 22.13 & 96.32 & 89.79 \\
15 & 1.5 & 0.4015 & 124.65 & 21.85 & 96.22 & 89.78 \\
\bottomrule
\end{tabular}
\end{minipage}
\end{table*}

\begin{table*}[ht]
\centering
\caption{OmniEraser Results: (Left: BenchHard / Right: RemovalBench)}
\setlength{\tabcolsep}{3pt} 

\begin{minipage}[t]{0.48\textwidth}
\centering
\begin{tabular}{l|lllll}
\toprule
CFG scale & LPIPS & FID & PSNR & CLIP & DINO \\
\midrule
1.0 & 0.6173 & 265.49 & 14.19 & 86.75 & 69.99 \\
2.0 & 0.6261 & 246.16 & 13.32 & 86.93 & 73.16 \\
\textbf{3.5} & \textbf{0.3663} & \textbf{77.76} & \textbf{21.15} & \textbf{97.25} & \textbf{93.91} \\
7.0 & 0.6390 & 230.97 & 13.83 & 88.36 & 74.87 \\
\bottomrule
\end{tabular}
\end{minipage}
\hfill
\begin{minipage}[t]{0.48\textwidth}
\centering
\begin{tabular}{l|lllll}
\toprule
Scale & LPIPS & FID & PSNR & CLIP & DINO \\
\midrule
1.0 & 0.6567 & 336.11 & 14.74 & 87.84 & 67.66 \\
2.0 & 0.6816 & 325.31 & 13.55 & 86.68 & 70.92 \\
\textbf{3.5} & \textbf{0.3774} & \textbf{70.88} & \textbf{22.65} & \textbf{97.53} & \textbf{94.16} \\
7.0 & 0.6940 & 309.08 & 13.88 & 88.14 & 72.85 \\
\bottomrule
\end{tabular}
\end{minipage}
\end{table*}

\begin{table*}[ht]
\centering
\setlength{\tabcolsep}{3pt} 
\caption{RORem Results: (Left: BenchHard / Right: RemovalBench)}
\begin{minipage}[t]{0.48\textwidth}
\centering
\begin{tabular}{l|lllll}
\toprule
CFG & LPIPS & FID & PSNR & CLIP & DINO \\
\midrule
\textbf{1} & \textbf{0.2429} & \textbf{64.15} & \textbf{22.42} & \textbf{98.01} & \textbf{94.61} \\
3 & 0.2436 & 64.10 & 22.40 & 98.01 & 94.58 \\
5 & 0.2438 & 64.12 & 22.38 & 98.00 & 94.55 \\
7 & 0.2449 & 64.81 & 22.34 & 97.99 & 94.47 \\
9 & 0.2464 & 65.58 & 22.30 & 97.98 & 94.38 \\
\bottomrule
\end{tabular}
\end{minipage}
\hfill
\begin{minipage}[t]{0.48\textwidth}
\centering
\begin{tabular}{l|lllll}
\toprule
CFG & LPIPS & FID & PSNR & CLIP & DINO \\
\midrule
1 & 0.2577 & 83.02 & 24.24 & 98.12 & 94.14 \\
3 & 0.2576 & 82.79 & 24.25 & 98.12 & 94.21 \\
\textbf{5} & \textbf{0.2577} & \textbf{82.66} & \textbf{24.26} & \textbf{98.13} & \textbf{94.22} \\
7 & 0.2578 & 82.90 & 24.27 & 98.13 & 94.22 \\
9 & 0.2587 & 83.12 & 24.26 & 98.12 & 94.19 \\
\bottomrule
\end{tabular}
\end{minipage}
\end{table*}

\begin{table*}[ht]
\centering
\begin{tabular}{lllll|lllll}
\toprule
cfg & \makecell{Rem. \\ guidance} & \makecell{Similarity \\ supp. steps} & \makecell{Similarity \\ supp. scale} & strength & LPIPS & FID & PSNR & CLIP & DINO \\
\midrule
1.0 & 7 & 5 & 0.3 & 0.8 & 0.2353 & 64.5465 & 22.4108 & 97.9236 & 94.2675 \\
1.0 & 7 & 5 & 0.3 & 0.9 & 0.2354 & 64.5416 & 22.4144 & 97.9399 & 94.3148 \\
1.0 & 7 & 5 & 0.3 & 1.0 & 0.2360 & 64.7933 & 22.3893 & 97.9089 & 94.2402 \\
1.0 & 7 & 5 & 0.5 & 0.8 & 0.2359 & 66.5109 & 22.2117 & 97.8727 & 94.2292 \\
\textbf{1.0} & \textbf{7} & \textbf{5} & \textbf{0.5} & \textbf{0.9} & \textbf{0.2358} & \textbf{62.9151} & \textbf{22.3571} & \textbf{97.9553} & \textbf{94.2937} \\
1.0 & 7 & 5 & 0.5 & 1.0 & 0.2360 & 64.5297 & 22.3854 & 97.9248 & 94.3316 \\
1.0 & 7 & 9 & 0.3 & 0.8 & 0.2352 & 65.7186 & 22.4169 & 97.8995 & 94.1419 \\
1.0 & 7 & 9 & 0.3 & 0.9 & 0.2353 & 66.3822 & 22.4285 & 97.8630 & 94.2854 \\
1.0 & 7 & 9 & 0.3 & 1.0 & 0.2358 & 63.4083 & 22.4153 & 97.9072 & 94.1839 \\
1.0 & 7 & 9 & 0.5 & 0.8 & 0.2354 & 65.6787 & 22.4187 & 97.9326 & 94.3263 \\
1.0 & 7 & 9 & 0.5 & 0.9 & 0.2353 & 65.1219 & 22.4715 & 97.9290 & 94.2453 \\
1.0 & 7 & 9 & 0.5 & 1.0 & 0.2357 & 65.8562 & 22.4509 & 97.8830 & 94.2129 \\
1.0 & 9 & 5 & 0.3 & 0.8 & 0.2354 & 65.8350 & 22.4294 & 97.8846 & 94.3280 \\
1.0 & 9 & 5 & 0.3 & 0.9 & 0.2354 & 64.9290 & 22.4649 & 97.9011 & 94.1800 \\
1.0 & 9 & 5 & 0.3 & 1.0 & 0.2358 & 64.5543 & 22.3940 & 97.9390 & 94.4061 \\
1.0 & 9 & 5 & 0.5 & 0.8 & 0.2355 & 66.9499 & 22.3768 & 97.9633 & 94.2039 \\
1.0 & 9 & 5 & 0.5 & 0.9 & 0.2357 & 64.9623 & 22.4054 & 97.9788 & 94.2309 \\
1.0 & 9 & 5 & 0.5 & 1.0 & 0.2358 & 64.7565 & 22.3793 & 97.9191 & 94.3760 \\
1.0 & 9 & 9 & 0.3 & 0.8 & 0.2354 & 65.6625 & 22.4099 & 97.8594 & 94.1816 \\
1.0 & 9 & 9 & 0.3 & 0.9 & 0.2356 & 64.9122 & 22.4341 & 97.8966 & 94.1760 \\
1.0 & 9 & 9 & 0.3 & 1.0 & 0.2358 & 64.5817 & 22.4314 & 97.8888 & 94.2095 \\
1.0 & 9 & 9 & 0.5 & 0.8 & 0.2890 & 68.0023 & 21.0889 & 96.4145 & 90.3991 \\
1.0 & 9 & 9 & 0.5 & 0.9 & 0.2356 & 64.7674 & 22.4499 & 97.9156 & 94.2936 \\
1.0 & 9 & 9 & 0.5 & 1.0 & 0.2358 & 64.9755 & 22.4365 & 97.8997 & 94.2588 \\
2.0 & 7 & 5 & 0.3 & 0.8 & 0.2353 & 64.5465 & 22.4108 & 97.9236 & 94.2675 \\
2.0 & 7 & 5 & 0.3 & 0.9 & 0.2354 & 64.5416 & 22.4144 & 97.9399 & 94.3148 \\
2.0 & 7 & 5 & 0.3 & 1.0 & 0.2360 & 64.7933 & 22.3893 & 97.9089 & 94.2402 \\
2.0 & 7 & 5 & 0.5 & 0.8 & 0.2359 & 66.5109 & 22.2117 & 97.8727 & 94.2292 \\
2.0 & 7 & 5 & 0.5 & 0.9 & 0.2358 & 62.9151 & 22.3571 & 97.9553 & 94.2937 \\
2.0 & 7 & 5 & 0.5 & 1.0 & 0.2360 & 64.5297 & 22.3854 & 97.9248 & 94.3316 \\
2.0 & 7 & 9 & 0.3 & 0.8 & 0.2352 & 65.7186 & 22.4169 & 97.8995 & 94.1419 \\
2.0 & 7 & 9 & 0.3 & 0.9 & 0.2353 & 66.3822 & 22.4285 & 97.8630 & 94.2854 \\
2.0 & 7 & 9 & 0.3 & 1.0 & 0.2358 & 63.4083 & 22.4153 & 97.9072 & 94.1839 \\
2.0 & 7 & 9 & 0.5 & 0.8 & 0.2354 & 65.6787 & 22.4187 & 97.9326 & 94.3263 \\
2.0 & 7 & 9 & 0.5 & 0.9 & 0.2353 & 65.1219 & 22.4715 & 97.9290 & 94.2453 \\
2.0 & 7 & 9 & 0.5 & 1.0 & 0.2357 & 65.8562 & 22.4509 & 97.8830 & 94.2129 \\
2.0 & 9 & 5 & 0.3 & 0.8 & 0.2354 & 65.8350 & 22.4294 & 97.8846 & 94.3280 \\
2.0 & 9 & 5 & 0.3 & 0.9 & 0.2354 & 64.9290 & 22.4649 & 97.9011 & 94.1800 \\
2.0 & 9 & 5 & 0.3 & 1.0 & 0.2358 & 64.5543 & 22.3940 & 97.9390 & 94.4061 \\
2.0 & 9 & 5 & 0.5 & 0.8 & 0.2355 & 66.9499 & 22.3768 & 97.9633 & 94.2039 \\
2.0 & 9 & 5 & 0.5 & 0.9 & 0.2357 & 64.9623 & 22.4054 & 97.9788 & 94.2309 \\
2.0 & 9 & 5 & 0.5 & 1.0 & 0.2358 & 64.7565 & 22.3793 & 97.9191 & 94.3760 \\
2.0 & 9 & 9 & 0.3 & 0.8 & 0.2354 & 65.6625 & 22.4099 & 97.8594 & 94.1816 \\
2.0 & 9 & 9 & 0.3 & 0.9 & 0.2356 & 64.9122 & 22.4341 & 97.8966 & 94.1760 \\
2.0 & 9 & 9 & 0.3 & 1.0 & 0.2358 & 64.5817 & 22.4314 & 97.8888 & 94.2095 \\
2.0 & 9 & 9 & 0.5 & 0.8 & 0.2354 & 65.4113 & 22.4455 & 97.9366 & 94.4160 \\
2.0 & 9 & 9 & 0.5 & 0.9 & 0.2356 & 64.7674 & 22.4499 & 97.9156 & 94.2936 \\
2.0 & 9 & 9 & 0.5 & 1.0 & 0.2358 & 64.9755 & 22.4365 & 97.8997 & 94.2588 \\
\bottomrule
\end{tabular}

\caption{Attentive Eraser (BenchHard)}
\end{table*}

\begin{table*}[ht]
\centering
\begin{tabular}{lllll|lllll}
\toprule
cfg & \makecell{Rem. \\ guidance} & \makecell{Similarity \\ supp. steps} & \makecell{Similarity \\ supp. scale} & strength & LPIPS & FID & PSNR & CLIP & DINO \\
\midrule
1.0 & 7 & 5 & 0.3 & 0.8 & 0.2559 & 79.5612 & 24.4449 & 98.0463 & 94.2709 \\
1.0 & 7 & 5 & 0.3 & 0.9 & 0.2557 & 77.4792 & 24.4695 & 98.0507 & 94.1880 \\
1.0 & 7 & 5 & 0.3 & 1.0 & 0.2557 & 78.5954 & 24.5227 & 98.0533 & 94.0470 \\
1.0 & 7 & 5 & 0.5 & 0.8 & 0.2561 & 78.7337 & 24.4234 & 98.0707 & 94.1455 \\
1.0 & 7 & 5 & 0.5 & 0.9 & 0.2556 & 78.2796 & 24.4913 & 98.0245 & 94.2425 \\
1.0 & 7 & 5 & 0.5 & 1.0 & 0.2557 & 77.6969 & 24.5258 & 98.0491 & 94.0396 \\
1.0 & 7 & 9 & 0.3 & 0.8 & 0.2560 & 79.4169 & 24.5148 & 98.0897 & 94.1993 \\
1.0 & 7 & 9 & 0.3 & 0.9 & 0.2559 & 77.9213 & 24.5126 & 98.0758 & 94.2251 \\
1.0 & 7 & 9 & 0.3 & 1.0 & 0.2558 & 78.3305 & 24.4982 & 98.0413 & 94.1997 \\
1.0 & 7 & 9 & 0.5 & 0.8 & 0.2558 & 80.0599 & 24.4904 & 98.0713 & 94.2864 \\
1.0 & 7 & 9 & 0.5 & 0.9 & 0.2557 & 78.0075 & 24.5081 & 98.0631 & 94.2688 \\
1.0 & 7 & 9 & 0.5 & 1.0 & 0.2557 & 76.3665 & 24.5213 & 98.0488 & 94.1518 \\
1.0 & 9 & 5 & 0.3 & 0.8 & 0.2558 & 79.8573 & 24.5034 & 98.0514 & 94.3472 \\
1.0 & 9 & 5 & 0.3 & 0.9 & 0.2558 & 78.5669 & 24.4793 & 98.0416 & 94.2727 \\
1.0 & 9 & 5 & 0.3 & 1.0 & 0.2557 & 77.1161 & 24.5108 & 98.0483 & 93.9702 \\
1.0 & 9 & 5 & 0.5 & 0.8 & 0.2558 & 79.4824 & 24.4835 & 98.0598 & 94.3191 \\
1.0 & 9 & 5 & 0.5 & 0.9 & 0.2557 & 77.6902 & 24.5004 & 98.0346 & 94.2263 \\
1.0 & 9 & 5 & 0.5 & 1.0 & 0.2557 & 77.3909 & 24.5164 & 98.0558 & 94.0791 \\
1.0 & 9 & 9 & 0.3 & 0.8 & 0.2562 & 78.8775 & 24.5164 & 98.0217 & 94.2556 \\
1.0 & 9 & 9 & 0.3 & 0.9 & 0.2560 & 78.7569 & 24.5121 & 98.0703 & 94.3393 \\
1.0 & 9 & 9 & 0.3 & 1.0 & 0.2558 & 77.8419 & 24.5052 & 98.0490 & 94.1366 \\
1.0 & 9 & 9 & 0.5 & 0.8 & 0.2562 & 79.0362 & 24.4625 & 98.0156 & 94.0172 \\
1.0 & 9 & 9 & 0.5 & 0.9 & 0.2558 & 77.7559 & 24.5118 & 98.0638 & 94.3270 \\
1.0 & 9 & 9 & 0.5 & 1.0 & 0.2557 & 77.1574 & 24.5194 & 98.0535 & 94.1307 \\
2.0 & 7 & 5 & 0.3 & 0.8 & 0.2559 & 79.5612 & 24.4449 & 98.0463 & 94.2709 \\
2.0 & 7 & 5 & 0.3 & 0.9 & 0.2557 & 77.4792 & 24.4695 & 98.0507 & 94.1880 \\
2.0 & 7 & 5 & 0.3 & 1.0 & 0.2557 & 78.5954 & 24.5227 & 98.0533 & 94.0470 \\
2.0 & 7 & 5 & 0.5 & 0.8 & 0.2561 & 78.7337 & 24.4234 & 98.0707 & 94.1455 \\
2.0 & 7 & 5 & 0.5 & 0.9 & 0.2556 & 78.2796 & 24.4913 & 98.0245 & 94.2425 \\
2.0 & 7 & 5 & 0.5 & 1.0 & 0.2557 & 77.6969 & 24.5258 & 98.0491 & 94.0396 \\
2.0 & 7 & 9 & 0.3 & 0.8 & 0.2560 & 79.4169 & 24.5148 & 98.0897 & 94.1993 \\
2.0 & 7 & 9 & 0.3 & 0.9 & 0.2559 & 77.9213 & 24.5126 & 98.0758 & 94.2251 \\
2.0 & 7 & 9 & 0.3 & 1.0 & 0.2558 & 78.3305 & 24.4982 & 98.0413 & 94.1997 \\
2.0 & 7 & 9 & 0.5 & 0.8 & 0.2558 & 80.0599 & 24.4904 & 98.0713 & 94.2864 \\
2.0 & 7 & 9 & 0.5 & 0.9 & 0.2557 & 78.0075 & 24.5081 & 98.0631 & 94.2688 \\
2.0 & 7 & 9 & 0.5 & 1.0 & 0.2557 & 76.3665 & 24.5213 & 98.0488 & 94.1518 \\
2.0 & 9 & 5 & 0.3 & 0.8 & 0.2558 & 79.8573 & 24.5034 & 98.0514 & 94.3472 \\
2.0 & 9 & 5 & 0.3 & 0.9 & 0.2558 & 78.5669 & 24.4793 & 98.0416 & 94.2727 \\
2.0 & 9 & 5 & 0.3 & 1.0 & 0.2557 & 77.1161 & 24.5108 & 98.0483 & 93.9702 \\
2.0 & 9 & 5 & 0.5 & 0.8 & 0.2558 & 79.4824 & 24.4835 & 98.0598 & 94.3191 \\
2.0 & 9 & 5 & 0.5 & 0.9 & 0.2557 & 77.6902 & 24.5004 & 98.0346 & 94.2263 \\
2.0 & 9 & 5 & 0.5 & 1.0 & 0.2557 & 77.3909 & 24.5164 & 98.0558 & 94.0791 \\
2.0 & 9 & 9 & 0.3 & 0.8 & 0.2562 & 78.8775 & 24.5164 & 98.0217 & 94.2556 \\
2.0 & 9 & 9 & 0.3 & 0.9 & 0.2560 & 78.7569 & 24.5121 & 98.0703 & 94.3393 \\
2.0 & 9 & 9 & 0.3 & 1.0 & 0.2558 & 77.8419 & 24.5052 & 98.0490 & 94.1366 \\
\textbf{2.0} & \textbf{9} & \textbf{9} & \textbf{0.5} & \textbf{0.8} & \textbf{0.2557} & \textbf{78.5961} & \textbf{24.5152} & \textbf{98.0720} & \textbf{94.3913} \\
2.0 & 9 & 9 & 0.5 & 0.9 & 0.2558 & 77.7559 & 24.5118 & 98.0638 & 94.3270 \\
2.0 & 9 & 9 & 0.5 & 1.0 & 0.2557 & 77.1574 & 24.5194 & 98.0535 & 94.1307 \\
\bottomrule
\end{tabular}

\caption{Attentive Eraser (RemovalBench)}
\end{table*}

\begin{table*}[ht]
\setlength{\tabcolsep}{3pt} 
\centering
\caption{Tuning of insertion algorithm on RORD validation set.}
\begin{minipage}[t]{0.32\textwidth}
\centering
\textbf{FreeCompose}\\[1pt]
\begin{tabular}{l|lllll}
\toprule
\makecell{DDS \\ weight} & LPIPS & FID & PSNR & CLIP & DINO \\
\midrule
1.0 & 0.3927 & 120.2 & 19.47 & 93.25 & 89.71 \\
\textbf{1.5} & \textbf{0.3901} & \textbf{119.9} & \textbf{19.49} & \textbf{93.30} & \textbf{89.76} \\
2.0 & 0.3904 & 120.0 & 19.48 & 93.29 & 89.70 \\
\bottomrule
\end{tabular}

\vspace{1em}
\textbf{Paint-by-Example}\\[1pt]
\begin{tabular}{l|lllll}
\toprule
\makecell{CFG \\ scale } & LPIPS & FID & PSNR & CLIP & DINO \\
\midrule
1.0 & 0.4190 & 146.0 & 17.54 & 91.87 & 82.59 \\
3.0 & 0.4183 & 135.1 & 17.47 & 92.98 & 84.99 \\
\textbf{5.0} & \textbf{0.4182} & \textbf{133.0} & \textbf{17.40} & \textbf{93.24} & \textbf{85.60} \\
7.0 & 0.4183 & 132.6 & 17.34 & 93.30 & 85.91 \\
9.0 & 0.4185 & 131.7 & 17.29 & 93.34 & 86.06 \\
\bottomrule
\end{tabular}
\end{minipage}
\hfill
\begin{minipage}[t]{0.64\textwidth}
\centering
\textbf{AnyDoor}\\[1pt]
\begin{tabular}{ll|lllll}
\toprule
\makecell{Guidance \\ scale} & \makecell{Control \\ scale} & LPIPS & FID & PSNR & CLIP & DINO \\
\midrule
1.0 & 0.4 & 0.4807 & 193.9 & 15.63 & 88.45 & 75.74 \\
1.0 & 0.8 & 0.4203 & 136.0 & 17.16 & 92.77 & 83.40 \\
1.0 & 1.0 & 0.4076 & 130.9 & 17.36 & 93.70 & 84.63 \\
1.0 & 1.2 & 0.4093 & 141.7 & 17.25 & 93.76 & 85.43 \\
1.0 & 1.5 & 0.4252 & 157.5 & 16.63 & 93.11 & 84.84 \\
1.0 & 2.0 & 0.4699 & 177.9 & 15.50 & 90.90 & 80.47 \\
4.5 & 0.4 & 0.4776 & 150.7 & 15.36 & 90.86 & 82.09 \\
4.5 & 0.8 & 0.4184 & 123.0 & 17.10 & 94.17 & 88.43 \\
\textbf{4.5} & \textbf{1.0} & \textbf{0.4065} & \textbf{130.2} & \textbf{17.25} & \textbf{94.54} & \textbf{89.56} \\
4.5 & 1.2 & 0.4224 & 122.2 & 16.67 & 94.23 & 89.28 \\
4.5 & 1.5 & 0.4430 & 125.0 & 15.88 & 93.81 & 89.02 \\
4.5 & 2.0 & 0.4710 & 138.4 & 15.12 & 92.97 & 87.64 \\
7.0 & 0.4 & 0.4710 & 164.6 & 15.46 & 91.45 & 83.03 \\
7.0 & 0.8 & 0.4248 & 125.2 & 16.90 & 94.00 & 88.38 \\
7.0 & 1.0 & 0.4136 & 132.2 & 16.99 & 94.34 & 89.55 \\
7.0 & 1.2 & 0.4309 & 126.2 & 16.42 & 93.93 & 89.29 \\
7.0 & 1.5 & 0.4414 & 145.7 & 15.88 & 93.52 & 89.23 \\
7.0 & 2.0 & 0.4835 & 133.6 & 14.78 & 92.52 & 87.22 \\
\bottomrule
\end{tabular}
\end{minipage}
\end{table*}

\begin{table*}[ht]
\centering
\begin{tabular}{lllll|lllll}
\toprule
$w_{content}$ & $w_{contrast}$ & $w_{id}$ & \makecell{Guidance \\ scale} & \makecell{Energy \\ scale} & LPIPS & FID & PSNR & CLIP & DINO \\
\midrule
3 & 0.1 & 1.8 & 7 & 0.9 & 0.3093 & 68.2051 & 18.2531 & 96.7788 & 92.8310 \\
3 & 0.1 & 1.8 & 7 & 1.5 & 0.3068 & 66.8024 & 18.2299 & 96.8619 & 93.0648 \\
3 & 0.1 & 2.5 & 7 & 0.9 & 0.3092 & 68.4845 & 18.2347 & 96.7950 & 92.7789 \\
3 & 0.1 & 2.5 & 7 & 1.5 & 0.3081 & 67.3742 & 18.1283 & 96.8288 & 92.9424 \\
3 & 0.2 & 1.8 & 7 & 0.9 & 0.3103 & 68.9174 & 18.2117 & 96.7664 & 92.7191 \\
3 & 0.2 & 1.8 & 7 & 1.5 & 0.3079 & 67.7704 & 18.1751 & 96.8291 & 92.9493 \\
3 & 0.2 & 2.5 & 7 & 0.9 & 0.3101 & 69.0386 & 18.2018 & 96.7840 & 92.6634 \\
3 & 0.2 & 2.5 & 7 & 1.5 & 0.3094 & 68.1056 & 18.0344 & 96.8164 & 92.8475 \\
3 & 0.3 & 1.8 & 7 & 0.9 & 0.3113 & 69.9344 & 18.1741 & 96.7384 & 92.6562 \\
3 & 0.3 & 1.8 & 7 & 1.5 & 0.3090 & 68.1417 & 18.1031 & 96.8240 & 92.8750 \\
3 & 0.3 & 2.5 & 7 & 0.9 & 0.3112 & 69.8445 & 18.1593 & 96.7535 & 92.6027 \\
3 & 0.3 & 2.5 & 7 & 1.5 & 0.3103 & 69.0084 & 17.9881 & 96.8104 & 92.8011 \\
6 & 0.1 & 1.8 & 7 & 0.9 & 0.3047 & 65.8831 & 18.3690 & 96.8665 & 93.2757 \\
\textbf{6} & \textbf{0.1} & \textbf{1.8} & \textbf{7} & \textbf{1.5} & \textbf{0.3038} & \textbf{64.8498} & \textbf{18.2943} & \textbf{96.9336} & \textbf{93.4044} \\
6 & 0.1 & 2.5 & 7 & 0.9 & 0.3047 & 66.1367 & 18.3477 & 96.8574 & 93.2554 \\
6 & 0.1 & 2.5 & 7 & 1.5 & 0.3058 & 66.0787 & 18.1736 & 96.8706 & 93.1908 \\
6 & 0.2 & 1.8 & 7 & 0.9 & 0.3051 & 66.2827 & 18.3445 & 96.8646 & 93.1979 \\
6 & 0.2 & 1.8 & 7 & 1.5 & 0.3047 & 65.7645 & 18.2499 & 96.9187 & 93.3180 \\
6 & 0.2 & 2.5 & 7 & 0.9 & 0.3052 & 66.4090 & 18.3242 & 96.8693 & 93.0798 \\
6 & 0.2 & 2.5 & 7 & 1.5 & 0.3059 & 66.6991 & 18.1355 & 96.8560 & 93.1302 \\
6 & 0.3 & 1.8 & 7 & 0.9 & 0.3065 & 73.7313 & 18.3854 & 96.8271 & 93.0749 \\
\bottomrule
\end{tabular}
\caption{DragonDiffusion}
\end{table*}


\end{document}